\def\year{2022}\relax
\documentclass[letterpaper]{article} 
\usepackage{aaai22_arxiv}  
\usepackage{times}  
\usepackage{helvet}  
\usepackage{courier}  
\usepackage[hyphens]{url}  
\usepackage{graphicx} 
\usepackage{natbib}  
\usepackage{caption} 
\usepackage{mathtools}
\usepackage{pifont}

\DeclareCaptionStyle{ruled}{labelfont=normalfont,labelsep=colon,strut=off} 
\usepackage{pxfonts}
\usepackage{txfonts}
\usepackage[ruled,vlined]{algorithm2e}

\usepackage{booktabs}
\usepackage{amsthm}

\theoremstyle{definition}
\newtheorem{defn}{Definition}
\newtheorem{prop}{Property}
\usepackage{newfloat}
\usepackage{listings}
\usepackage{xcolor}

\title{On Causally Disentangled Representations}

\author {
    Abbavaram Gowtham Reddy, 
    Benin Godfrey L, 
    Vineeth N Balasubramanian
}
\affiliations {
    Indian Institute of Technology Hyderabad, India\\
    cs19resch11002@iith.ac.in, benin.godfrey@cse.iith.ac.in, vineethnb@iith.ac.in  
}

\begin{document}
\maketitle
\begin{abstract}
Representation learners that disentangle factors of variation have already proven to be important in addressing various real world concerns such as fairness and interpretability. Initially consisting of unsupervised models with independence assumptions, more recently, weak supervision and correlated features have been explored, but without a causal view of the generative process. In contrast, we work under the regime of a causal generative process where generative factors are either independent or can be potentially confounded by a set of observed or unobserved confounders. We present an analysis of disentangled representations through the notion of disentangled causal process. We motivate the need for new metrics and datasets to study causal disentanglement and propose two evaluation metrics and a dataset. We show that our metrics capture the desiderata of disentangled causal process. 
Finally, we perform an empirical study on state of the art disentangled representation learners using our metrics and dataset to evaluate them from causal perspective.
\end{abstract}

\section{Introduction}
Humans implicitly tend to use causal reasoning while learning and explaining real-world concepts. Deep learning models, however, are considered to be \textit{black-box}~\cite{lipton2018mythos} and also \textit{correlational}, thus, we cannot directly rely on their decisions in safety-critical domains such as medicine, defence, aerospace, etc. Consequently, there has been a surge in using the ideas of causality to improve the learning and explanation capabilities of deep learning models in recent years~\cite{causal_explanation_black_box,suter2019robustly,goyal2019explaining, goyal2019counterfactual, chattopadhyay2019neural, janzing2019causal, zmigrod2019counterfactual, pitis2020counterfactual, zhu2019causal,towards_causal_repr}. Deep learning models that learn the underlying causal structures in data not only avoid this problem of learning purely correlational input-output relationships, but also help in providing causal explanations. In this work, we choose disentangled representation learning as a tool to study the usefulness of applying causality in machine learning.

Disentangled representation learning~\cite{repr_lear, towards_causal_repr} aims to identify the underlying independent generative factors of variation given an observed data distribution, and is an important problem to address given its applications to generalization~\cite{montero2021the}, data generation~\cite{disent_generation}, explainability~\cite{exp_exp}, fairness~\cite{creager2019flexibly}, etc. The generative processes underlying observational data often contain complex interactions among generative factors. Treating such interactions as \textit{independent causal mechanisms}~\cite{peters2017elements} is essential to many real-world applications including the development of learning algorithms that learn transferable mechanisms from one domain to another~\cite{towards_causal_repr}.

The study of disentanglement in unsupervised settings, with independence assumptions on the generative factors, has been the dominant topic of study for some time in recent literature~\cite{Higgins2017betaVAELB,kumar2017variational,kim2018disentangling,chen2018isolating}. Considering the limitations of unsupervised disentanglement~\cite{locatello2019challenging} and potentially unrealistic nature of the independence assumptions, a few semi-supervised and weakly supervised disentanglement methods have also been developed more recently ~\cite{locatello2020weakly,Chen_Batmanghelich_2020,dittadi2021on,disent_corr}.
None of the abovementioned methods, however, take a causal view on the underlying data generative process while studying disentanglement. 
We study disentanglement from a causal perspective in this work, grounding ourselves on the very little work along this direction~\cite{suter2019robustly}. Since causal generative processes can be complex with arbitrary depth and width in their graphical representations, we restrict ourselves to two-level causal generative processes of the form shown in Figure~\ref{fig:dcp} as these, by itself, can model many real-world settings with confounding~\cite{pearl2009causality}, and have not been studied before either in the context of disentanglement or representation learning. We then also study how well-known latent variable models -- e.g., $\beta$-VAE~\cite{Higgins2017betaVAELB} -- perform disentanglement in the presence of confounders.

To this end, based on the definition of a \textit{disentangled causal process} by \cite{suter2019robustly}, we look at three essential properties of causal disentanglement and propose evaluation metrics that are grounded on the principles of causality to study the level of causal disentanglement achieved by a generative latent variable model. 
The analysis in \cite{suter2019robustly} focused on a metric for interventional robustness, and was studied w.r.t. the encoder of a latent variable model, which limits us to operating on only the interventional distribution of the encoder output. We instead extend the definition of \textit{disentangled causal process} to both the encoder and generator of a latent variable model. Studying disentanglement from the generator's perspective allows us to study the \textit{counterfactual distribution} of the generator output along with the \textit{interventional distribution} of the encoder output, thus enabling us to propose newer evaluation metrics to study causally disentangled representations.

Going further, given the limitations in existing datasets for study of causally disentangled representations -- especially their realism, natural confounding, and complexity -- we introduce a new realistic image dataset, CANDLE, whose generation follows a two-level causal generative process with confounders, considering our focus in this work. The CANDLE dataset, along with the procedures for its creation, are made publicly available at \url{https://causal-disentanglement.github.io/IITH-CANDLE/}. We also perform empirical studies on popular latent variable models to understand their ability to causally disentangle the underlying generative process using our metrics, our dataset as well as on existing datasets in this regard. We summarize our key contributions below:
\begin{itemize}
    \item We undertake a study of causal perspectives to disentanglement, and go beyond existing work to capture the generative process of latent variable representation learning models, and thus study interventional and counterfactual goodness.
    \item We present two new evaluation metrics to study disentangled representation learning that are consequences of the properties of causally disentangled latent variable models.
    \item We introduce a new image-based dataset that includes known causal generative factors as well as confounders to help study and improve deep generative latent variable models from a causal perspective.
    \item We perform empirical studies on various well-known latent variable models in this regard, analyze their performance from a causal perspective and also show how a small degree of weak supervision can help improve causally disentangled representation learning.
\end{itemize}

\section{Related Work}

\noindent \textbf{Capturing the Generative Process.}
Evidently, the underlying causal generative process has an impact on understanding the level of disentanglement achieved by a model. For e.g., if two generative factors are correlated or confounded by external factors, existing models find it difficult to disentangle the underlying generative factors~\cite{disent_corr,dittadi2021on}. Much of the existing disentanglement literature relies on the assumption that generative factors are independent of each other~\cite{Higgins2017betaVAELB,kim2018disentangling,chen2018isolating}, and do not consider a causal view to the generating process. Recently,~\cite{suter2019robustly} presented a causal view to the generative process but focused on studying interventional robustness. We build on this work to present the desiderata of latent variable models to achieve causal disentanglement.

\noindent \textbf{Disentanglement in Representation Learning.}
Disentangled representation learning has been largely studied in unsupervised generative models in the last few years~\cite{Higgins2017betaVAELB,kumar2017variational,kim2018disentangling,chen2018isolating}. These methods essentially assume that the learned generative (or latent) factors are independent. Recently, semi-supervised and weakly supervised methods have been proposed~\cite{locatello2020weakly,Chen_Batmanghelich_2020,dittadi2021on,disent_corr} to achieve better disentanglement between the latent variables. However, these methods do not consider or study the alignment of such a learned disentangled representation to the causal generative model. Models that consider causal relationships among input features and learn structural causal models~\cite{pearl2009causality} in latent space have been proposed of late~\cite{yang2020causalvae,kocaoglu2018causalgan}; however, such efforts have been far and few between, and evaluating the extent of causal disentanglement has not been the objective of such methods. 

\noindent \textbf{Evaluation Metrics for Disentanglement.}
Existing work on learning disentangled representations using latent variable models have largely developed their own metrics to evaluate the extent of disentanglement, including the BetaVAE metric~\cite{Higgins2017betaVAELB}, FactorVAE metric~\cite{kim2018disentangling}, Mutual Information Gap~\cite{chen2018isolating}, Modularity~\cite{ridgeway2018learning}, DCI Disentanglement~\cite{eastwood2018framework}, and the SAP Score~\cite{kumar2017variational}. One important drawback of these metrics is that the possible effects of confounding in a generative process are not considered. Confounding is a critical aspect of real-world generative processes where the relationship between two variables can in turn depend on other variables (called \textit{confounding variables} or \textit{confounders}, see Figure \ref{fig:dcp}), which could either be observed or unobserved. Confounders are the reasons to observe spurious correlations among generative factors in observational data. This is one of the primary challenges in studying causal effects, and requires careful consideration when evaluating disentangled representations. The first causal effort in this direction was the Interventional Robustness Score (IRS) developed by ~\cite{suter2019robustly}, which however relies exclusively on the learned latent space to evaluate disentangled representations. The IRS metric allows for presence of confounders in the data generating process, but does not make an effort to differentiate them in the learned latent variable space (e.g., two generative factors that are highly correlated can still be encoded by a single latent factor, which can be limiting). We empirically observe that one can get a good IRS score with very little training (please see Appendix) but at the cost of bad reconstructions, i.e. the IRS metric does not capture the goodness of the disentangled latent variables in generating useful data. Good reconstructions and thus good counterfactual generations are equally important in our quest to achieve deep learning models that learn causal generative factors. Our proposed evaluation metrics address this important issue by penalizing the latent variables that are confounded and by quantitatively evaluating the generated counterfactuals.

\noindent \textbf{Image Datasets for Study of Disentanglement.}
Image datasets that are studied in disentangled representation learning include dSprites~\cite{dsprites17}, smallNORB~\cite{lecun2004learning}, 3Dshapes~\cite{3dshapes18}, cars3D~\cite{cars3d}, MPI3D~\cite{gondal2019transfer}, Falcor3D, and Isaac3D~\cite{falcor3d}. These datasets, which are mostly synthetic, are generated based on a causal graph in which all factors of variation are assumed to be independent, and the causal graph is largely one-level. We introduce a realistic image dataset that involves two-level causal graphs with semantically relevant confounders to study various disentanglement methods using our proposed metrics. More details including comparisons with existing datasets are presented in Section~\ref{sec:dataset} and in the Appendix.


\section{Disentangled Causal Process}
\begin{figure}
		\centering
        \includegraphics[width=0.6\linewidth]{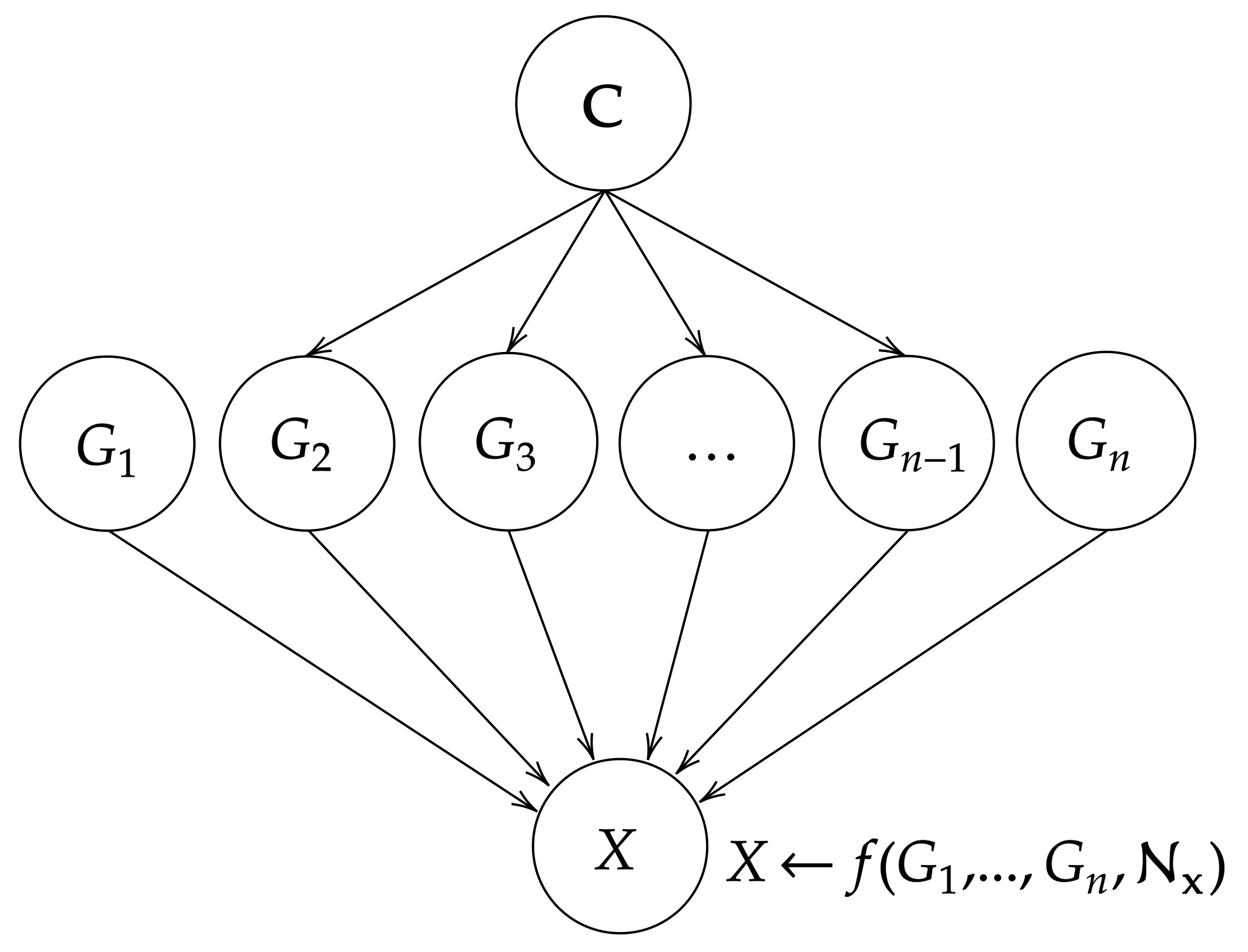}
		\caption{A causal process for generating $X$ with generative factors $\{G_1,\dots,G_n\}$ and confounders $\mathbf{C}$}
		\label{fig:dcp}
\end{figure}

We work under the regime of causal generative processes of the form shown in Figure~\ref{fig:dcp} where a set of generative factors $\mathbf{G}=\{G_1, G_2,\dots, G_n\}$ are independent by nature but can potentially be confounded by a set of confounders ($\mathbf{C}$). 
To this end, we begin by stating the definition of \textit{disentangled causal process}~\cite{suter2019robustly} below. 
\label{dcp}
\begin{defn}
\label{def:disentangled_causal_process}
(Disentangled Causal Process \cite{suter2019robustly}). When a set of generative factors $\mathbf{G}=\{G_1,\dots,G_n\}$ do not causally influence each other (i.e., $G_i\not \rightarrow G_j$) but can be confounded by a set of confounders $\mathbf{C}=\{C_1,\dots,C_l\}$, a causal model for $X$ with generative factors $\mathbf{G}$ is said to be disentangled if and only if it can be described by a \textit{structural causal model}~\cite{pearl2009causality} of the form: 
\begin{align*}
    C_j &\leftarrow \mathcal{N}_{c_j}; j\in\{1,\dots,l\}\\
    G_i &\leftarrow g_i(PA^{C}_i, \mathcal{N}_{G_i}); i\in\{1,\dots,n\}\\
    X &\leftarrow f(G_1,\dots,G_n, \mathcal{N}_x)
\end{align*}
where $f,g_i$ are independent causal mechanisms, $PA^C_i\subseteq \{C_1,\dots,C_l\}$ represents the parents of $G_i$ and $\mathcal{N}_{c_j}, \mathcal{N}_{G_i}, \mathcal{N}_{x}$ are independent noise variables. 
\end{defn}
We now examine an essential property (Property~\ref{prop1}) of such a disentangled causal process and we extend it to latent variable models to be able to propose new evaluation metrics for causal disentanglement. 
\begin{prop}
\label{prop1}
In a disentangled causal process of type shown in Figure~\ref{fig:dcp}, $G_i$ does not causally influence $G_j, i\neq j$ because any intervention on $G_i$ will remove incoming edges from $\mathbf{C}$ and $X$ is a collider in the path $G_i \rightarrow X \leftarrow G_j$~\cite{pearl2009causality}. As a consequence, $G_i$ will not have any causal effect on $X$ via $G_j$ and all the causal effect of $G_i$ on $X$ is via the directed edge $G_i\rightarrow X$~\cite{suter2019robustly}. 
\end{prop}

We now translate Property \ref{prop1} to deep generative latent variable models. Considering their well-known use in disentanglement literature, we focus on Variational Auto-Encoders (VAEs) for this purpose. A latent variable model $\mathcal{M}\ (e, g, p_X)$ with an encoder $e$, generator $g$ and a data distribution $p_X$, assumes a prior $p(Z)$ on the latent space, and a generator $g$ (often a deep neural network) is  parametrized as $p_{\theta}(X|Z)$. We then approximate the posterior $p(Z|X)$ using a variational distribution $q_{\phi}(Z|X)$ parametrized by another deep neural network, called the encoder $e$. The prior is usually assumed to be an isotropic Gaussian~\cite{kingma2013auto,rezende14} and the model is trained on $x \sim p_X$ by maximizing the log-likelihood of reconstruction and minimizing the difference between the prior and approximated posterior. This leads to a set of generative factors $\mathbf{G}$ encoded as a set of latent dimensions $\mathbf{Z}=\{Z_1,\dots, Z_m\}$. Specifically the latent variable model captures each generative factor $G_i$ as a set of latent dimensions $\mathbf{Z}_I \subseteq \mathbf{Z}$ ($\mathbf{Z}$ indexed by a set of indices $I$). Ideally, one would want $I$ to be a singleton set so each generative factor has a unique latent variable learned by the model, but it is also possible for such a model to encode $G_i$ into more than one latent dimension (e.g., an angle can be encoded as $\sin{\theta}, \cos{\theta}$ in two different latent dimensions~\cite{ridgeway2018learning}). 
In order for latent variable models to view the generator $g$ as a causal mechanism to generate observations $\hat{x}$, $\mathbf{Z}$ acts now as a proxy for the true generative factors $\mathbf{G}$ (we use $x$ for an instance of a random variable $X$, $\hat{x}$ hence denotes the reconstruction of $x$ obtained through the generator $g$). 

As a consequence of Property~\ref{prop1}, any latent variable model $\mathcal{M}$ should satisfy the following two properties to achieve causal disentanglement.
\begin{prop}
\label{prop2}
If a latent variable model $\mathcal{M}\ (e,g,p_X)$ disentangles a causal process of type shown in Figure~\ref{fig:dcp}, and the encoder $e$ learns a latent space $\mathbf{Z}$ such that each generative factor $G_i$ is  mapped to a unique $\mathbf{Z}_I$ (unique $\mathbf{Z}_I$ refers to the scenario: $\mathbf{Z}_I \cap \mathbf{Z}_J = \emptyset, I\neq J, |I|,|J| \geq 0$ where $\mathbf{Z}_I$ is responsible for another generative factor $G_j$), then the generator $g$ is a disentangled causal mechanism that models the underlying generative process.
\end{prop}
Property~\ref{prop2} is similar to \textit{encoder disentanglement} in~\cite{shu2019weakly} but we view it in terms of the generator than the encoder. Property~\ref{prop2} essentially boils down to learning a one-to-one mapping between each $G_i$ and $\mathbf{Z}_I$, i.e. when two data points $x_1, x_2$ differ in only one generative factor $G_i$, one should observe a change only in $\mathbf{Z}_I$ when generating $\hat{x}_1, \hat{x}_2$.
\begin{prop}
\label{prop3}
In a latent variable model $\mathcal{M}\ (e,g,p_X)$ that disentangles a causal process of type shown in Figure~\ref{fig:dcp}, the only causal feature of $\hat{x}$ w.r.t. generative factor $G_i$ is $\mathbf{Z}_I \forall i$.
\end{prop}

We now propose two evaluation metrics in the next section that are consequences of Properties~\ref{prop2} and~\ref{prop3}. To study the disentanglement of a causal process of the type shown in Figure~\ref{fig:dcp}, we need datasets that reflect the generative process, and we hence introduce one in Section~\ref{sec:dataset} which offers several advantages such as realism, semantic confounding and complex backgrounds over existing datasets in addition to being generated from a two-level causal graph with confounders.

\section{Evaluation Metrics}
\label{sec:metrics}
For causal disentanglement, the encoder $e$ of a model $\mathcal{M}\ (e, g, p_X)$ should learn the \textit{mapping} from $G_i$ to $\mathbf{Z}_I$ without any influence from confounding in the data distribution $p_X$. (This would be equivalent to marginalizing over the confounder while computing direct causal effect between two variables.) If a model is able to map each $G_i$ to a unique $\mathbf{Z}_I$, we say that the learned latent space $\mathbf{Z}$ is unconfounded. We call this property as \textit{Unconfoundedness} $(UC)$. $UC$ captures the essentials of Property~\ref{prop2} as it relies on the mapping between $G_i$ and $\mathbf{Z}_I$. 

Secondly, when the latent space is unconfounded, a counterfactual instance of $x$ w.r.t. generative factor $G_i$, $x^{cf}_I$ (i.e., the counterfactual of $x$ with change in only $G_i$) can be generated by intervening on the latents of $x$ corresponding to $G_i$, $\mathbf{Z}_I^x$ and any change in the latent dimensions of $\mathbf{Z}$ that are not responsible for generating $G_i$, i.e. $\mathbf{Z}_{\setminus I}^x$, should have no influence on the generated counterfactual instance $x^{cf}_I$ w.r.t. generative factor $G_i$. We call this property as \textit{Counterfactual Generativeness} $(CG)$. To explain with an example, consider an image of an ball in a certain background. The $CG$ metric emphasises the fact that \textit{``intervening on the latents corresponding to the background should only change the background and intervening on the latents corresponding to texture or shape of the ball should not change the background"}. Thus, $CG$ follows from Property~\ref{prop3} as it is based on the fact that only causal effect on $x^{cf}_I$ w.r.t. generative factor $G_i$ is from $\mathbf{Z}_I^x$. We now formally define the two metrics. 

\subsection{Unconfoundedness (UC) Metric}
\begin{figure*}
        \centering
          \includegraphics[width=0.7\textwidth]{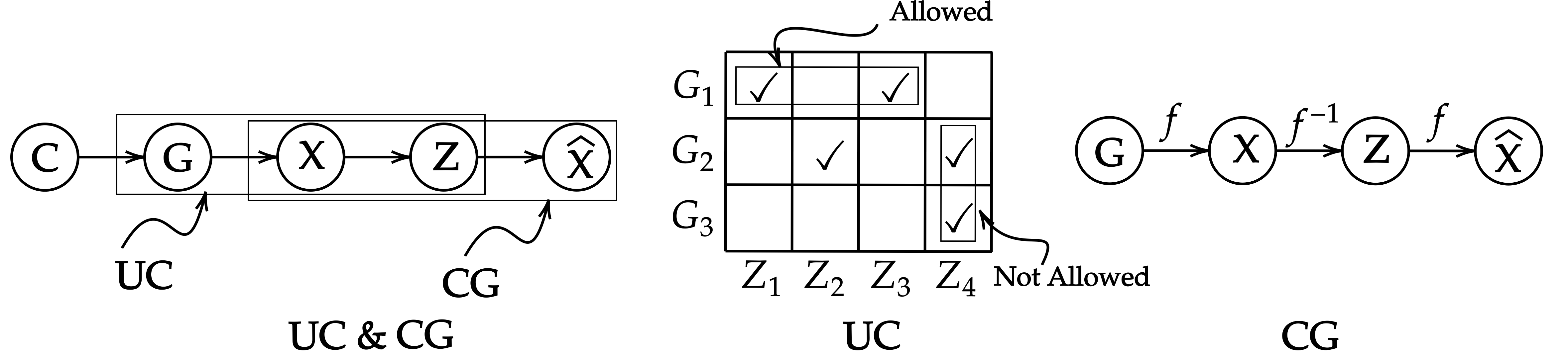}
          \caption{\footnotesize \textbf{Left}: $UC$ metric relates $\mathbf{G}$, $X$ and $\mathbf{Z}$. $CG$ metric relates $x$, $\mathbf{Z}$ and $\hat{x}$. \textbf{Center}: According to $UC$ metric, in a model
          $\mathcal{M}, G_1$ is allowed to be captured by $Z_1, Z_3$ but it is not allowed for $Z_4$ to capture both $G_2, G_3$ (this would suggest confounding). \textbf{Right}: Generative factors $\mathbf{G}$ generate image $x$ through an unknown causal mechanism $f$, our goal in learning a disentangled representation is to learn $f^{-1}$ and hence $f$ that transforms observation $x$ into latent dimensions $\mathbf{Z}$ and latent dimensions to reconstruction $\hat{x}$.}
        \label{fig:ucncg}
\end{figure*}

The $UC$ metric evaluates how well distinct generative factors $G_i$ are captured by distinct sets of latent dimensions $\mathbf{Z}_I$ with no overlap (Figure \ref{fig:ucncg}). If a model encodes the underlying generative factor $G_i$ of an instance $x$ as a set of latent dimensions $\mathbf{Z}_I^x$, we define $UC$ measure as: 
\begin{equation}
\label{eq:uc}
   UC \coloneqq 1 - \mathbb{E}_{x\sim p_X}\left [\frac{1}{S} \sum_{I,J} \frac{|\mathbf{Z}_I^x \cap \mathbf{Z}_J^x|}{|\mathbf{Z}_I^x \cup \mathbf{Z}_J^x|}\right ]
\end{equation}
where $S= \binom{n}{2}$ is the number of pairs of generative factors $(G_i,G_j), i\ne j$. We are in effect, finding the \textit{Jaccard similarity coefficient} among all possible pairs of latent variables corresponding to different $(G_i,G_j)$ to know how each pair of $(G_i,G_j)$ are captured by unconfounded latent dimensions. To find correspondences between $\mathbf{Z}_I$ and $G_i$, we can use any existing metrics like~\cite{suter2019robustly, chen2018isolating} but we use the IRS measure~\cite{suter2019robustly} as it works on principles of interventions and is grounded on the properties of a disentangled causal process. For each generative factor $G_i$, IRS finds latents $Z_I$ that are robust to interventions to $G_j; j\neq i$. If all generative factors are disentangled into distinct sets of latent factors, we get a $UC$ score of $1$. If all generative factors share the same set of latent factors, we get a $UC$ score of $0$. This definition of $UC$ metric can be generalized to also check for unconfoundedness of multiple generative factors at a time.

Metrics closest to $UC$ are $MIG$~\cite{chen2018isolating} and $DCI$~\cite{eastwood2018framework}. Even though $MIG$ penalizes non-axis aligned representations, it does not consider the case of multiple generative factors having the same latent representation, and hence may not capture unconfoundedness in a true sense. The Disentanglement$(D)$ score in $DCI$ uses correlation-based models to predict $G_i$ given $Z$, and is hence not causal.

\subsection{Counterfactual Generativeness (CG) Metric}
When a latent variable model $\mathcal{M}$ achieves \textit{unconfoundedness}, we can perform interventions on any specific $\mathbf{Z}_I$ to generate counterfactual instances without any confounding effect. That is, the generator $g$ is able to generate counterfactual instances in a flexible and controlled manner. We call this \textit{counterfactual generativeness}. In latent variable models that work on image datasets, to the best of our knowledge, this is the first effort to use generated images to \textit{quantitatively} evaluate the level of disentanglement. To define $CG$ metric mathematically, we need the notion of Average and Individual Causal Effect, which we provide below.
\begin{defn}
\label{def:ace}
(Average Causal Effect). The Average Causal Effect ($ACE$) of a random variable $Z$ on a random variable $X$ for a treatment $do(Z)=\alpha$ with reference to a baseline treatment $do(Z)=\alpha^*$ is defined as $ACE_{do(Z=\alpha)}^X \coloneqq \mathbb{E}[X|do(Z=\alpha)] - \mathbb{E}[X|do(Z=\alpha^*)]$.
\end{defn}
Individual Causal Effect $(ICE)$ can be defined similar to Definition \ref{def:ace} by replacing the expectation with probability as $ICE_{do(Z=\alpha)}^x \coloneqq p[x|do(Z=\alpha)] - p[x|do(Z=\alpha^*)]$. Perfect disentanglement makes the generative model satisfy the positivity assumption~\cite{hernan2019causal} 
and allows us to approximate $ACE$ with mean of $ICE$s taken over the dataset. 
Based on the above definitions, our \textit{counterfactual generativeness} ($CG$) metric is defined as:
\begin{equation}
    \label{cg_eqn}
    CG \coloneqq \mathbb{E}_I \big[| ACE^{X^{cf}_{I}}_{\mathbf{Z}_{I}^X} - ACE^{X^{cf}_{\setminus I}}_{\mathbf{Z}_{\setminus I}^X}|\big]
\end{equation}
$ACE^{X^{cf}_{I}}_{\mathbf{Z}_{I}^X}$ and $ACE^{X^{cf}_{\setminus I}}_{\mathbf{Z}_{\setminus I}^X}$ are defined to be the average causal effects of $\mathbf{Z}_I^x$ and $\mathbf{Z}_{\setminus I}^x$ on the respective counterfactual quantities $x^{cf}_{I}$ and $x^{cf}_{\setminus I}$ (recall that $I \subset \{1,2,\dots,m\}$ denotes the set of indices among the latent factors learned in the model that correspond to the $G_i^{\text{th}}$ generative factor).
So, the $CG$ metric calculates the normalized sum of differences of average causal effects of $\mathbf{Z}_I^x$ and $\mathbf{Z}_{\setminus I}^x$ on the generated counterfactual quantities w.r.t. $G_i$ (recall that for causal disentanglement, only $\mathbf{Z}_I^x$ should have causal effect on $x^{cf}_I$ w.r.t. generative factor $G_i$; recall the ball, background example). Since counterfactual outcomes with respect to a model can be generated through interventions, we approximate $ACE$ with the average of $ICE$s taken over the empirical distribution $p_X$. The practical version of the $CG$ metric is hence:
\begin{equation}
\label{eq:cg}
\begin{aligned}
    CG &\coloneqq \mathbb{E}_I \left[| ACE^{X^{cf}_{I}}_{\mathbf{Z}_{I}^X} - ACE^{X^{cf}_{\setminus I}}_{\mathbf{Z}_{\setminus I}^X}|\right]  
    \approx \mathbb{E}_I \left[|\mathbb{E}_{x\sim p_x} [ICE^{x^{cf}_I}_{\mathbf{Z}_I^x} - ICE^{x^{cf}_{\setminus I}}_{\mathbf{Z}_{\setminus I}^x}]|\right] \\
    &\approx \frac{1}{n} \left[|\frac{1}{L} [ICE^{x^{cf}_I}_{\mathbf{Z}_I^x} - ICE^{x^{cf}_{\setminus I}}_{\mathbf{Z}_{\setminus I}^x}]| \right]\\
\end{aligned}
\end{equation}
where $L$ is the size of the dataset. Definition \ref{def:ace} holds for \textit{real} random variables, but in latent variable models, $x^{cf}_I$ is an image on which there is no clear way of defining causal effect of latents. Extending the notations, let $G_{ik}^x$ represent the $\textit{k}^{\text{th}}$ value taken by $\textit{i}^{\text{th}}$ generative factor for a specific image $x$ (e.g., if $\textit{i}=$ shape of an object, then $\textit{k}=$ cone). For this work, we define $ICE^{x^{cf}_I}_{\mathbf{Z}_I^x}$ to be the difference in prediction probability (of a pre-trained classifier) of $G_{ik}^x$ given the counterfactual image $x^{cf}_I$ generated when $do(\mathbf{Z}_I^x = \mathbf{Z}_I^x)$ (i.e. no change in latents of current instance) and when $do(\mathbf{Z}_I^x = baseline(\mathbf{Z}_I^x))$. Mathematically,
\begin{equation}
\label{eq:ice1}
\begin{aligned}
    ICE^{x^{cf}_I}_{\mathbf{Z}_I^x} &\coloneqq |p(G_{ik}^x|x^{cf}_I, do(\mathbf{Z}_I^x=\mathbf{Z}_I^x))\\
    &- p(G_{ik}^x|x^{cf}_I, do(\mathbf{Z}_I^x=baseline(\mathbf{Z}_I^x))|
\end{aligned}
\end{equation}
\begin{equation}
\label{eq:ice2}
\begin{aligned}
        ICE^{x^{cf}_{\setminus I}}_{\mathbf{Z}_{\setminus I}^x} &\coloneqq |p(G_{ik}^x|x^{cf}_{\setminus I}, do(\mathbf{Z}_{\setminus I}^x=\mathbf{Z}_{\setminus I}^x)) \\
        &- p(G_{ik}^x|x^{cf}_{\setminus I}, do(\mathbf{Z}_{\setminus I}^x=baseline(\mathbf{Z}_{\setminus I}^x))|\\
\end{aligned}
\end{equation}
We use $baseline(\mathbf{Z}_I^x)$ as the latent dimensions that are maximally deviated from the current latent values $\mathbf{Z}_I^x$ (taken over the dataset) to ensure that we get a reasonably different image from the current image $x$ w.r.t. generative factor $G_i$.
$baseline(\mathbf{Z}_I^x)$ can be $0$ or $\mathbb{E}_{x\sim p_X}(\mathbf{Z}_I^x)$ depending on the dataset and application. In the ideal scenario, Equation \ref{eq:ice1} is expected to output $1$ because $\mathbf{Z}_I^x$ is the only causal feature of $G^x_{ik}$. Equation \ref{eq:ice2} is expected to output $0$ because $\mathbf{Z}_{\setminus I}^x$ is not causally responsible for generating $G^x_{ik}$. Now it is easy to see that, for causal disentanglement, $CG$ score in Equation \ref{eq:cg} is $1$; and for poor disentanglement, $CG$ score is $0$. The proposed $UC$ and $CG$ metrics can also be used irrespective of presence of confounders in the data generating process. The algorithms detailing the implementation of $UC$ and $CG$ metrics are provided in the Appendix~\ref{additional_details_on_metrics}.

\section{Dataset}
\label{sec:dataset}
\begin{figure}
		\centering
        \includegraphics[width=0.55\linewidth]{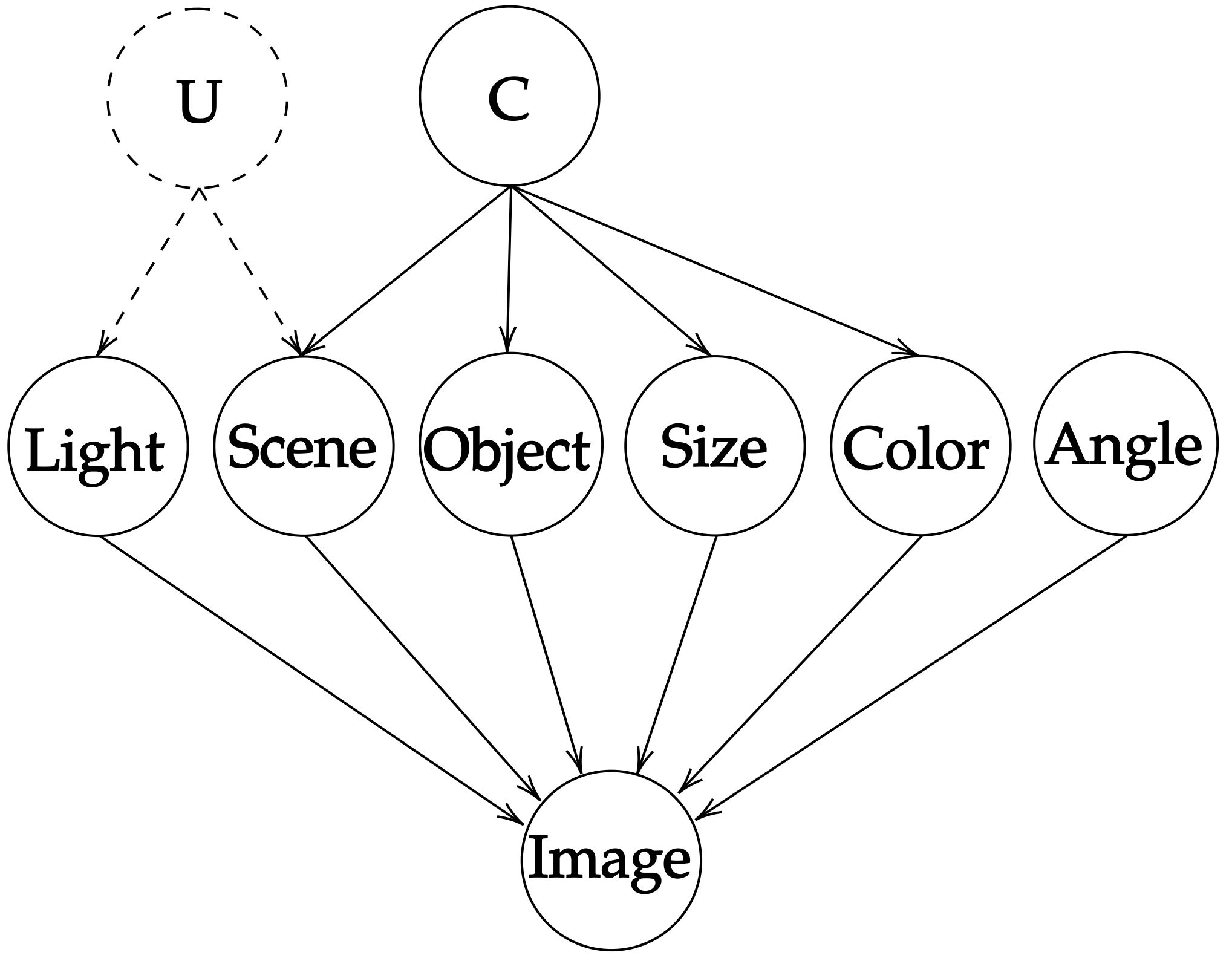}
		\caption{Image generating process of CANDLE}
		\label{fig:datagenerator_confounding}
\end{figure}

\begin{figure}
   \centering
   \includegraphics[width=0.36\textwidth]{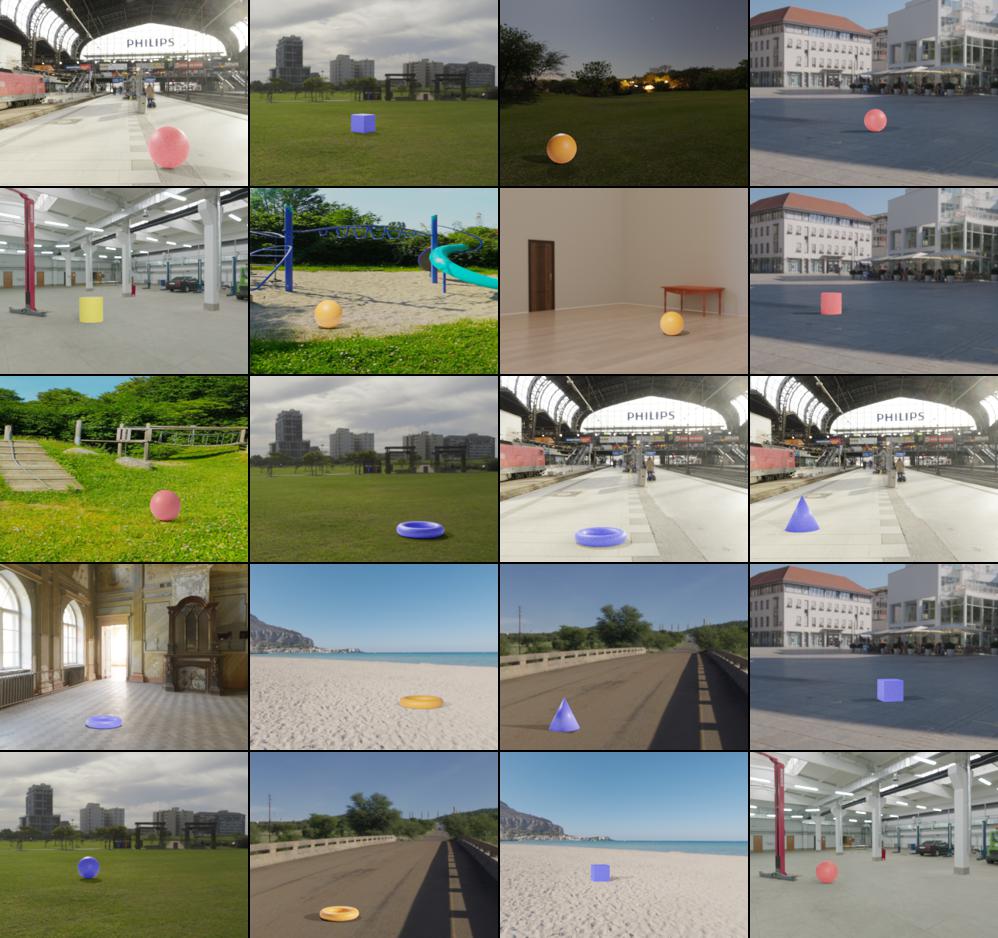}
   \caption{\footnotesize Sample images from CANDLE. Different objects appear in different colors, shapes, rotations and in different backgrounds respecting the causal graph in Figure \ref{fig:datagenerator_confounding}}
   \label{fig:sample_images}
\end{figure}

To study causally disentangled representations, we introduce an image dataset called CANDLE (\underline{C}ausal \underline{AN}alysis in \underline{D}isentang\underline{L}ed r\underline{E}presentations) with 6 data generating factors along with both observed and unobserved confounders. Its generation follows the causal directed acyclic graph shown in Figure \ref{fig:datagenerator_confounding} which resembles our setting of causal graphs introduced in Figure~\ref{fig:dcp}.
During generation, the \textit{Image} has influences from confounders $\mathbf{U}$ (unobserved), and $\mathbf{C}$ (observed) through intermediate generative factors such as \textit{Object} and \textit{size}. It contains observed confounding in the form of semantic constraints such as overly large objects not being in indoor scenes (full list in Appendix). Unobserved confounding shows up in the interaction between the artificial light source and the scene's natural lighting conditions as it interacts with the foreground object producing shadows. Another source of subtle confounding in the dataset is how the location of the object and its size are confounded by \textit{depth}, where a larger object that is farther-off and a smaller object nearby occupy the same pixel real-estate in the image, explored in \cite{disent_corr}. Sample images from CANDLE are shown in figure~\ref{fig:sample_images} and a comparison with existing datasets used commonly in disentangled representation learning is provided in the Appendix~\ref{appendix:comparison_of_datasets} where we also highlight the important features that are unique to CANDLE compared to existing datasets.

\paragraph{Dataset Creation.}
CANDLE is generated using Blender~\cite{blender}, a free and open-source 3D computer graphics suite which allows for manipulating background high-dynamic range images (HDRI images) and adding foreground elements that inherit the natural light of the background. Foreground elements also naturally cast shadows and this greatly increases the realism of the dataset while allowing for it to be simulated for ease of creation. Since high-quality HDRI images are easy to obtain, it allows for multiple realistic backgrounds, unlike the plain colors in dSprites~\cite{dsprites17} or colored strips in MPI3D~\cite{gondal2019transfer}. Having complex backgrounds and the position of the foreground object varying between images adds another level of complexity while modeling the dataset for any downstream task. Having specific objects of interest in representation learning tasks puts more responsibility on the models being learned on the dataset to reconstruct images that do not leave out small objects in the reconstruction. To aid such reconstructions, bounding boxes of foreground objects are included in CANDLE's metadata. To further help with realism, we ensure that the capturing camera was not kept stationary and produced a fair amount of random jitter. At every stage, the dataset is made in such a way that extensions to it by adding objects or modifying the background scene is trivial (see Appendix~\ref{details_on_dataset}).

CANDLE consists of $12,546$ images as $320\times240$ images and corresponding JSON files containing the factors of variation of each image (samples included in supplementary material). Recent works in disentangled representation learning have focused on identifying causal relationships in latent representations and the causal effects of latent representations on outcome variables~\cite{yang2020causalvae, chattopadhyay2019neural}, which our dataset can readily support due to availability of the required ground truth. Another use case of CANDLE would be in counterfactual generation algorithms~\cite{chang2018explaining, goyal2019explaining, ilse2020diva}, which we leave for future work. 

\paragraph{Details of Factors of Variation.}
Background scenes of CANDLE are panoramic HDRI images of $4000\times2000$ resolution for accurate reproduction of the scene's lights on the object. Foreground objects are placed on the floor (without complete occlusion to guarantee presence of every label in the image). Objects are sized for semantic correctness in relation to the background (e.g., juxtaposing a very large cube and a building is unrealistic). Care is taken to make sure that significant overlap between objects and the background is eliminated. An artificial light source is added to the scene which also casts shadows in 3 positions - left, middle (overhead) and right. This is an unobserved confounding variable in the sense that it could conflict with the scene's illumination. The light source is kept invariant across all objects in the image i.e., the light's position is the same irrespective of other object variables. The rotations of foreground objects are in the vertical axis. This variable is specifically chosen as it has visible differences in a subset of objects but may be interpreted as noise in the rest. For more details on CANDLE, please see Appendix~\ref{details_on_dataset}. We empirically observe that when the object of interest is small in the image and the image contains significant variations in the background scene, unlike on datasets such as MPI3D~\cite{gondal2019transfer} where foreground object is small but background is black/plain, reconstructions by standard latent variable models tend to not retain the foreground objects. One can use high multiplicative factors for the reconstruction term in the learned objective function, but this leads to bad latent representations \cite{kim2018disentangling}. We show how the bounding box information provided in CANDLE's metadata is used as weak supervision to solve this problem partially in Section~\ref{sec:imp_uc}. 

\section{Learning Disentangled Representations using Weak Supervision}
\label{sec:imp_uc}
We now provide a simple methodology to improve over existing models that learn disentangled representations by using the bounding box-level supervision information in CANDLE. Since there is a known trade-off between reconstruction quality and disentanglement in VAE-based models\cite{kim2018disentangling}, instead of giving high weightage to reconstruction quality during training at the cost of worse disentanglement, we hypothesize that paying more attention to the quality of reconstructions of specific foreground objects whose bounding box is known provides a more favorable trade-off between reconstructions and disentanglement. We improve the existing semi-supervised Factor-VAE\cite{kim2018disentangling} loss with an additional loss term that weights regions in the bounding box higher than others to aid in better reconstructions of foreground objects. We call this method \textit{Semi-Supervised Factor-VAE with additional Bounding Box supervision} or \textit{SS-FVAE-BB}. Our loss function w.r.t. dataset $\mathcal{D}=\{x_i\}_{i=1}^L$ is given by:
\begin{equation}
    \mathcal{L}_{SS-FVAE-BB} = \mathcal{L}_{(Factor-VAE)} +  \lambda \sum_{i=1}^L ||x_i\odot w_i - \hat{x}_i\odot w_i||^2_2
\end{equation}
where $w_i\in \{0,1\}^{320\times240\times3}$ is an indicator tensor with $1$s in the region of the bounding box and $0$s elsewhere, $\lambda$ is a hyperparameter and $\odot$ is the Hadamard (elementwise) product. Our experimental results (Table \ref{tab:causaldataset_experiments}) show that the proposed method improves $UC$ score while matching the best $CG$ score achieved by state-of-the-art models. \textit{SS-FVAE-BB} can also be used with the datasets without bounding box information by using any segmentation techniques that highlight the objects of interest in the images.

\section{Experimental Results}
To study causal disentanglement, we performed experiments on well-known unsupervised disentanglement methods as well as their corresponding semi-supervised variants: $\beta$-VAE~\cite{Higgins2017betaVAELB}, $\beta$-TCVAE~\cite{chen2018isolating}, DIP-VAE~\cite{kumar2017variational}, and Factor-VAE~\cite{kim2018disentangling} using the proposed dataset and evaluation metrics. We also included studies on other existing datasets --  dSprites, MPI3D-Toy, and a synthetic toy dataset with extreme confounding -- for completeness of analysis and comparison. The learned models are compared using $IRS, DCI(D), UC$ and $CG$ metrics. We use the open-source disentanglement library~\cite{locatello2019challenging} for training models. Semi-supervision is provided by using labels for 10\% of data points. Additional details on the experimental setup and qualitative results are provided in the Appendix~\ref{details_on_experiments}. In the results below, $\rho$ refers to the number of latent dimensions that we choose to attribute for each generative factor.
\begin{table}[h]
\begin{center}
\footnotesize
\scalebox{0.9}{
\begin{tabular}{lcccccc}
\toprule
Model  & $IRS$&$DCI$ & $UC$&$CG$& $UC$&$CG$\\
&&($D$)&$\rho=5$&$\rho=5$&$\rho=7$&$\rho=7$\\
\midrule
$\beta$-VAE &0.85& 0.18& 0.11& 0.24& 0.08& 0.22\\
$\beta$-TCVAE &0.82&0.10&0.11&0.25&0.08&0.25\\
DIP-VAE&0.33&0.08&0.11&0.21&0.15&0.22\\
Factor-VAE &\textbf{0.88}&0.15&0.13&0.26&0.08&\textbf{0.28}\\
SS-$\beta$-VAE&0.74&\textbf{0.18}&0.11&\textbf{0.28}&0.08&0.19\\
SS-$\beta$-TCVAE&0.68&0.17&0.11&0.23&0.08&0.19\\
SS-DIP-VAE&0.35&0.08&0.11&0.22&0.15&0.22\\
SS-Factor-VAE&0.61&0.16&0.24&\textbf{0.28}&0.14&0.22\\
\midrule
\textbf{SS-FVAE-BB}&0.61&0.13&\textbf{0.27}&\textbf{0.28}&\textbf{0.18}&\textbf{0.28}\\
\bottomrule
\end{tabular}
}
\end{center}
\vspace{-6pt}
\caption{Comparison of $IRS, DCI(D), UC$ and $CG$ metrics on CANDLE dataset}
\vspace{-6pt}
\label{tab:causaldataset_experiments}
\end{table}

\begin{table}[h]
\begin{center}
\footnotesize
\scalebox{0.9}{
\begin{tabular}{lcccccc}
\toprule
Model  & $IRS$ &$DCI$& $ UC $ & $CG$& $UC$&$CG$\\
&&($D$)&$\rho=1$&$\rho=1$&$\rho=2$&$\rho=2$\\
 \midrule 
$\beta$-VAE     &0.49 & 0.16 & 0.70      &0.12   &0.46   &0.10\\
$\beta$-TCVAE   &\textbf{0.78} & 0.43 &\textbf{0.90}&\textbf{0.19}   &0.60   &\textbf{0.19}\\
DIP-VAE         &0.12 & 0.03 &\textbf{0.90} &0.04  &0.60   &0.03\\
Factor-VAE      &0.44 & 0.13 &\textbf{0.90} &0.07   &0.60   &0.06\\
SS-$\beta$-VAE  &0.52 & 0.23 &\textbf{0.90} &0.17&0.60&0.17\\
SS-$\beta$-TCVAE&0.72 & \textbf{0.50} &\textbf{0.90}  &0.18   &\textbf{0.67}   &0.18\\
SS-DIP-VAE      &0.20 & 0.04 &0.40  &0.08   &0.13   &0.06\\
SS-Factor-VAE   &0.47 & 0.19 &\textbf{0.90}      &0.15   &0.33   &0.14\\
\bottomrule
\end{tabular}
}
\end{center}
\vspace{-6pt}
\caption{\footnotesize Comparison of $IRS, DCI(D), UC, CG$ metrics on dSprites}
\vspace{-6pt}
\label{tab:dsprites_experiment}
\end{table}
\vspace{-6pt}
\begin{table}[h]
\begin{center}
\footnotesize
\scalebox{0.9}{
\begin{tabular}{lcccccc}
\toprule
Model  & $IRS$ &$DCI$& $ UC $ & $CG$& $UC$&$CG$\\
&&($D$)&$\rho=1$&$\rho=1$&$\rho=2$&$\rho=2$\\
 \midrule 
$\beta$-VAE     &0.57&0.23&0.52&0.10&0.34&0.12\\
$\beta$-TCVAE   &0.57&0.22&0.52&0.12&0.35&0.14\\
DIP-VAE         &0.22&0.23&0.28&0.10&0.19&0.14\\
Factor-VAE      &0.52&\textbf{0.34}&0.71&\textbf{0.14}&0.47&\textbf{0.16} \\
SS-$\beta$-VAE  &0.60&0.28&\textbf{0.80}&0.10&\textbf{0.67}&0.09\\
SS-$\beta$-TCVAE&\textbf{0.64}&0.26&\textbf{0.80}&0.09&\textbf{0.67}&0.15\\
SS-DIP-VAE      &0.35&0.25&0.52&0.10&0.34&0.11\\
SS-Factor-VAE   &0.56&0.30&\textbf{0.80}&0.12&\textbf{0.67}&0.14\\
\bottomrule
\end{tabular}
}
\end{center}
\vspace{-6pt}
\caption{\footnotesize Comparison of $IRS, DCI(D), UC, CG$ metrics on MPI3D-Toy dataset}
\label{tab:mpi3dtoy_experiment}
\vspace{-6pt}
\end{table}

\noindent \textbf{Results on CANDLE:}
Table \ref{tab:causaldataset_experiments} shows the results of different performance metrics, including the proposed $UC$ and $CG$ metrics, when the considered generative models are learned on the CANDLE dataset (the '\textit{SS-}' prefix refers to the 'Semi-Supervised' variants). 
The table shows low $UC$ and $CG$ scores in general, motivating the need for better disentanglement methods. Owing to the complex background, models find it difficult to reconstruct foreground objects during the learning process which causes the learned latent dimensions corresponding to foreground objects difficult to identify. Changing the value of $\rho$ has its consequences. We observe higher (but not high enough for good disentanglement) $UC$ scores when $\rho=5$. However, when $\rho=7$, we observe low $UC$ scores because multiple latent dimensions are confounded. Owing to the complex background, models learn to reconstruct images with little to no information about the foreground object which also leads to low $CG$ scores. Much of the observed $CG$ score can be attributed to the \textit{scene} factor because scenes are reconstructed well (see Appendix). Introducing weak supervision in the training of the generative model using our proposed method \textit{SS-FVAE-BB} with $\lambda=2$ improves $UC$ score without compromising $CG$ score. 

\noindent \textbf{Results on dSprites \& MPI3D-Toy:} For completeness of analysis, we conducted experiments on training the abovementioned generative models on existing datasets with no confounding like dSprites \& MPI3D (Tables \ref{tab:dsprites_experiment}, \ref{tab:mpi3dtoy_experiment}). The $UC$ and $CG$ metrics can be used to evaluate models under this setting too. As we are training models on full datasets without any observable confounding effect, we observe high $UC$ scores when $\rho=1$. However, when $\rho=2$, results start to show limitations of existing models to disentangle completely. 
Additional results on confounded versions of dSprites and MPI3D-Toy datasets, as well as on a synthetic toy dataset with confounding that we created for purposes of analysis, are deferred to the Appendix owing to space constraints.
The results in general show that there is no single model that outperforms w.r.t. all the metrics, which shows the importance of datasets like CANDLE and evaluation metrics, such as $UC$ and $CG$ scores developed using the principles of causality, to uncover sources of bias that were not considered previously.
\section{Conclusions}
A causal view of disentangled representations is important for learning trustworthy and transferable mechanisms from one domain to another . We build on the very little work along this direction by analysing the properties of causal disentanglement in latent variable models. We propose two evaluation metrics and a dataset which are used to uncover the causal disentanglement in existing disentanglement methods. We also improved over existing models by introducing a simple weakly supervised disentanglement method. We hope that newer machine learning models benefit from our metrics and dataset in developing causally disentangled representation learners.
\paragraph{Acknowledgements.}
We are grateful to the Ministry of Education, India for the financial support of this work through the Prime Minister's Research Fellowship (PMRF) and UAY programs. This work has also been partly supported by Honeywell and a Google Research Scholar Award, whom we are thankful to. We thank the anonymous reviewers for their valuable feedback that helped improve the presentation of this work.
\bibliography{aaai22}
\clearpage

\appendix
\section{Appendix}
This appendix we provide the following details:
\begin{enumerate}
\setlength\itemsep{-0.1em}
    \item Additional details of $UC, CG$ metrics
    \begin{itemize}
    \setlength\itemsep{0em}
        \item Algorithms for implementation of $UC, CG$ metrics
        \item Time complexity of $UC, CG$ metrics
        \item Analysis of $UC$ metric
    \end{itemize}
    \item Additional details of CANDLE dataset
    \begin{itemize}
    \setlength\itemsep{0em}
        \item Details of factors of variation
        \item Metadata structure
        \item Observed confounding in CANDLE
        \item Details on the image rendering process
        \item Details on extensibility
        \item Some more sample images
        \item Comparison with popular datasets in disentanglement literature 
    \end{itemize}
    \item Additional experimental results
    \begin{itemize}
    \setlength\itemsep{0em}
        \item Additional details on experimental setup
        \item Experiments on a synthetic dataset with full confounding
        \item Experiments on confounded dSprites \& confounded MPI3D-Toy datasets
        \item Counterfactual images generated while computing $CG$ metric for CANDLE dataset
        \item Qualitative results of experiments on CANDLE and synthetic datasets
    \end{itemize}
    \item Assets and Licensing
\end{enumerate}

\section{Additional Details of Evaluation Metrics}
\label{additional_details_on_metrics}
\paragraph{Algorithms for Implementation of $UC, CG$ Metrics.}
Algorithms \ref{algo:metric_uc} and \ref{algo:metric_cg} show the steps to implement the computation of $UC$ and $CG$ metrics respectively.
\begin{algorithm}
\footnotesize
\SetAlgoLined
\textbf{Inputs:} Generative factors $\mathbf{G}$, learned latent dimensions $\mathbf{Z}$, IRS function\;
\KwResult{$UC$ metric}
\textbf{Initialize:} $T = 0$; $n = |\mathbf{G}|$ \;
\For{$i=0;\ i<n;\ i++$}{
$\mathbf{Z}_I$ = IRS($G_i$)\;
\For{$j=i+1;\ i<n-1;\ j++$}{
$\mathbf{Z}_J$ = IRS($G_j$)\;
$T = T + \frac{|\mathbf{Z}_I \cap \mathbf{Z}_J|}{|\mathbf{Z}_I \cup \mathbf{Z}_J|}$\;
}
}
$UC = 1 - \frac{2\times T}{n\times(n-1)}$ \;
return $UC$\;
 \caption{Unconfoundedness Metric}
 \label{algo:metric_uc}
\end{algorithm}

\begin{algorithm}
\footnotesize
\SetAlgoLined
\textbf{Inputs:} Generative factors $\mathbf{G}$, learned latent dimensions $\mathbf{Z}$, IRS function, dataset $\mathcal{D}$, trained generative model $g$\;
\KwResult{$CG$ metric}
\textbf{Initialize:} $CG = 0, n = |\mathbf{G}|, L= |\mathcal{D}|, ACE = 0$\;
\For{$i=0;\ i<n;\ i++$}{
$\mathbf{Z}_I$ = IRS($G_i$)\;
}

\For{$j=0;\ j<L;\ j++$}{
$x = x_j$\;
$x^{cf1}_I = g(\mathbf{Z}^x|do(\mathbf{Z}_I^x=\mathbf{Z}_I^x))$\;
$x^{cf2}_I = g(\mathbf{Z}^x|do(\mathbf{Z}_I^x=baseline(\mathbf{Z}_I^x))$\;
$ICE^{x^{cf}_I}_{\mathbf{Z}_I^x} = |P(G_{ik}^x|x^{cf1}_I) - P(G_{ik}^x|x^{cf2}_I)|$\;

$x^{cf1}_{\setminus I} = g(\mathbf{Z}^x|do(\mathbf{Z}_{\setminus I}^x=\mathbf{Z}_{\setminus I}^x))$\;
$x^{cf2}_{\setminus I} = g(\mathbf{Z}^x|do(\mathbf{Z}_{\setminus I}^x=baseline(\mathbf{Z}_{\setminus I}^x))$\;
$ICE^{x^{cf}_I}_{\mathbf{Z}_{\setminus I}^x} = |P(G_{ik}^x|x^{cf1}_{\setminus I}) - P(G_{ik}^x|x^{cf2}_{\setminus I})|$\;

$ACE = ACE + |ICE^{x^{cf}_I}_{\mathbf{Z}_I^x} - ICE^{x^{cf}_I}_{\mathbf{Z}_{\setminus I}^x}|$\;
}
$CG = \frac{ACE}{L}$\;
return $CG$\;
 \caption{Counterfactual Generativeness  Metric}
 \label{algo:metric_cg}
\end{algorithm}
\begin{figure*}
  \begin{center}
    \includegraphics[width=0.7\textwidth]{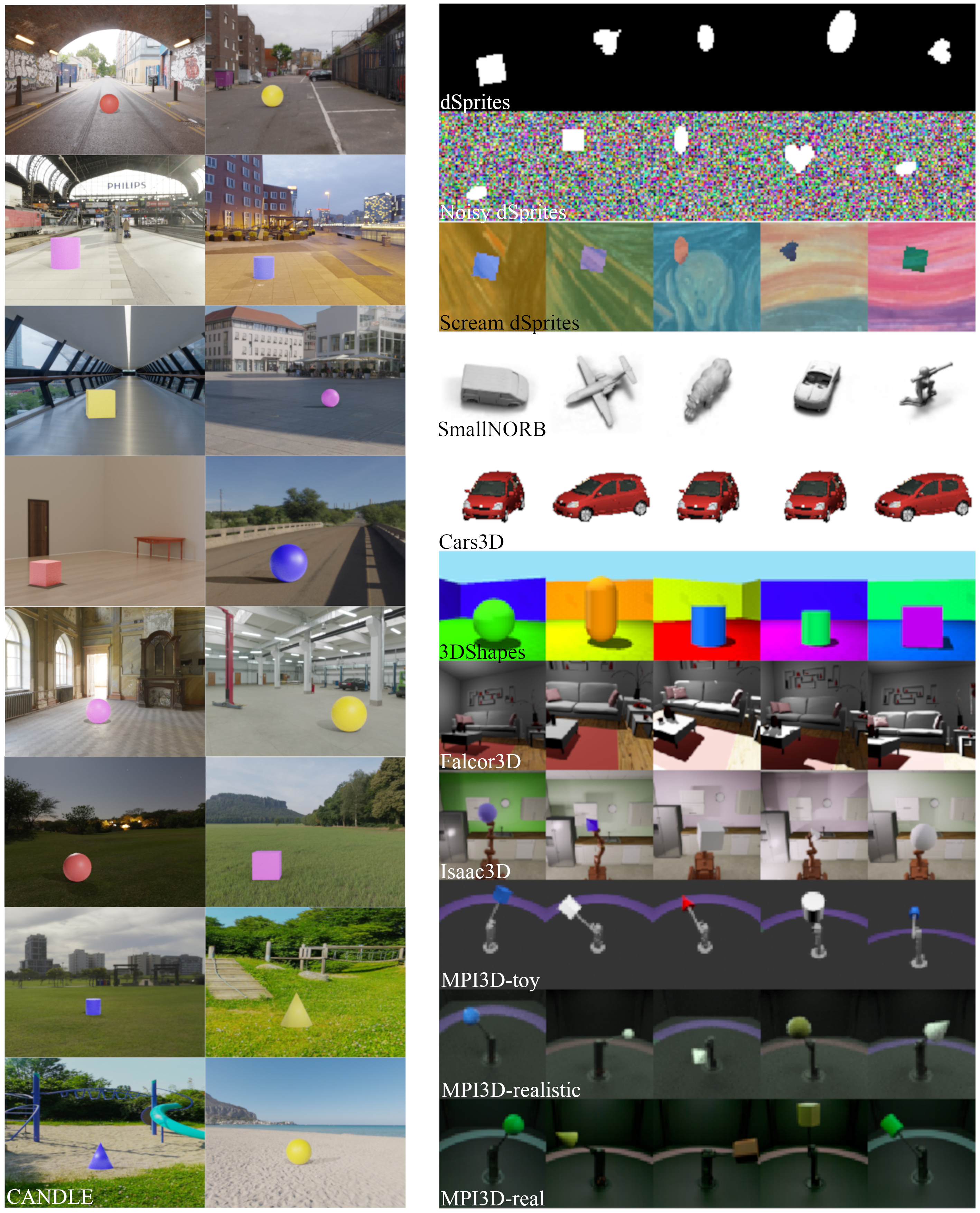}
  \end{center}
  \caption{Comparison of sample images from various datasets. Datasets \textit{(Left)}: CANDLE \textit{(Right: from top to bottom)}: dSprites, Noisy dSprites, Scream dSprites, SmallNORB, Cars3D, 3DShapes, Falcor3D, Isaac3D, MPI3D-toy, MPI3D-realistic, MPI3D-real. CANDLE is the only dataset with real and complex backgrounds developed using 2 level causal graph.}
  \label{appendix:comparison_of_datasets}
\end{figure*}
\begin{table*}
\begin{center}
\scalebox{0.9}{
\begin{tabular}{lcccccccc}
\toprule
Dataset& Depth of Underlying & 3D & Realistic & Presence of & Foreground Object  & Complex & Confounders \\
& Causal Graph & & & Foreground Object & Not Centered & Background &\\ 
\midrule
dSprites&1      &\ding{55}&\ding{55}&\ding{51}&\ding{51}&\ding{55} &\ding{55}\\
Noisy dsprites&1&\ding{55}&\ding{55}&\ding{51}&\ding{51}&\ding{55}&\ding{55}\\
Scream dsprites&1&\ding{55}&\ding{55}&\ding{51}&\ding{51}&\ding{55}&\ding{55}\\
SmallNORB&1    &\ding{51}&\ding{55}&\ding{51}&\ding{55}&\ding{55}&\ding{55}\\
Cars3D&1       &\ding{51}&\ding{55}&\ding{51}&\ding{55}&\ding{55}&\ding{55}\\
3Dshapes&1     &\ding{51}&\ding{55}&\ding{51}&\ding{55}&\ding{55}&\ding{55}\\
Falcor3D&1     &\ding{51}&\ding{55}&\ding{55}&\ding{55}&\ding{51}&\ding{55}\\
Isaac3D&1      &\ding{51}&\ding{55}&\ding{51}&\ding{51}&\ding{51}&\ding{55}\\
MPI3D-toy&1    &\ding{51}&\ding{55}&\ding{51}&\ding{51}&\ding{55}&\ding{55}\\
MPI3D-realistic&1&\ding{51}&\ding{51}&\ding{51}&\ding{51}&\ding{55}&\ding{55}\\
MPI3D-real&1   &\ding{51}&\ding{51}&\ding{51}&\ding{51}&\ding{55}&\ding{55}\\
Imagenet-C&N/A&\ding{51}&\ding{51}&\ding{51}&\ding{51}&\ding{51}&N/A\\
CIFAR-10/100-C&N/A&\ding{51}&\ding{51}&\ding{51}&\ding{51}&\ding{51}&N/A\\
Colored-MNIST&N/A&\ding{51}&\ding{55}&\ding{51}&\ding{55}&\ding{55}&N/A\\
PACS&N/A&N/A&N/A&\ding{51}&\ding{51}&N/A&N/A\\
Office-Home&N/A&N/A&N/A&\ding{51}&\ding{51}&N/A&N/A\\
\textbf{CANDLE}&\textbf{2}  &\ding{51}&\ding{51}&\ding{51}&\ding{51}&\ding{51}&\ding{51}\\
\bottomrule
\end{tabular}
}
\end{center}
\caption{Comparison of CANDLE with various existing datasets used in disentanglement and out of distribution (OOD) generalization tasks. CANDLE stands out after comparing with existing datasets along various dimensions. N/A: Not Applicable.}
\label{tab:datasets}
\end{table*}
\paragraph{Time Complexity of \textit{UC, CG} Metrics.}
\label{sec:complexity}
While evaluating $UC$ and $CG$ metrics (Eqns. 1 and 3 of main paper), the latents $\mathbf{Z}_I$ corresponding to each $G_i$ are obtained using IRS~\cite{suter2019robustly}, which was shown to have $\mathcal{O}(L)$ complexity~\cite{suter2019robustly}, where $L$ is the dataset size. Once we obtain $\mathbf{Z}_I$ corresponding to $G_i$, evaluation of the expression for $UC$ takes $\mathcal{O}(n^2)$ time where $n$ is the number of generative factors (usually a small number). To evaluate $CG$, we need to evaluate the prediction probabilities of $G^x_{ik}$ given the generated counterfactual image $x^{cf}_I$. Since the classifier is pre-trained, we can evaluate $CG$ for a single image $x$ using two forward passes through the network for each generative factor. The $CG$ algorithm hence runs in $\mathcal{O}(L\times n)$ time. Since $n$ (number of generative factors) is usually a small number, time complexity of $UC$ and $CG$ metrics is approximately linear in $L$.

\paragraph{Analysis of \textit{UC} Metric.}
$UC$ metric (Eqn. 1 of main paper) produces results that are densely distributed near $1$ because of the way Jaccard similarity behaves. This can be seen with the help of the following example. Consider the case where we have 2 generative factors and 6 latent dimensions. Assume that we attribute 3 latent dimensions for each generative factor (i.e., $\rho=3$). Now, let latents corresponding to the two generative factors be $\{1,2,3\}$ and $\{2,3,6\}$ respectively. In this case, $UC$ measure outputs 0.5 even though there is a significant overlap in the two sets. This effect of $UC$ scores hovering closer to 1 is however not a problem when we compare methods, since the relative differences between the values are more important here, not the absolute values.

\section{Additional Details of CANDLE Dataset}
\label{details_on_dataset}
\paragraph{Details of Factors of Variation.}
Table \ref{tab:generating_factors} shows the list of values taken by the generative factors: \textit{light, scene, object, size, color} and \textit{angle} that are part of the causal generative process (Fig. \ref{fig:datagenerator_confounding_appendix}) of CANDLE dataset.

\paragraph{Metadata of CANDLE.}

Figure \ref{fig:image_4150} and the adjoining image show a sample image and corresponding ground truth information of that image in JSON format. Size takes three values: $small(1.5),\ medium(2),\ and\ large(2.5)$. Bounding boxes (``\texttt{bounds}'') contain the bottom-left and top-right $(x,y)$ coordinates of the foreground object (i.e., \textit{object} factor in Fig.~\ref{fig:datagenerator_confounding}) in the image. 

Beyond learning unsupervised generative models on the dataset, having access to meta data allows parsing and querying over the ground-truth for specific variants of the factors as required to pair-up for weak supervision algorithms~\cite{locatello2020weakly, chen2020weakly} that are less susceptible to inductive biases~\cite{locatello2019challenging}. Paired images that differ in one or few generative factors (e.g., two images that differ in only background) as supervision to learn disentanglement has been explored recently~\cite{locatello2020weakly, chen2020weakly}. In addition to such pairing, weak supervision for representation learning models is also available in the dataset in other ways~\cite{shu2019weakly}. Match pairing, where we pair images with the same value for particular factors, can be done by querying for a subset containing the same value for the factors and pairing them up. Rank pairing -- which is match pairing with a ranking variable between the paired images based on a factor's value -- can also be done by querying and comparing values for these factors before pairing. The metadata thus allows current and future learning models to use the provided ground truth for learning and evaluation as required. 
\begin{table*}
\begin{minipage}{0.45\linewidth}
		\centering
		\footnotesize
        \begin{tabular}{cl}
        \toprule
        \textbf{Generative} &\textbf{Possible Values}\\
        \textbf{Factor}&\\
        \midrule
        Light& Left, Middle, Right\\
        \midrule
        &Indoor, Playground, Outdoor, Bridge,\\
        &City Square, Hall, Grassland, Garage,\\
        Scene&Street, Beach, Station,\\
        &Tunnel, Moonlit Grass, Dusk City,\\
        &Skywalk, Garden\\
        \midrule
        Object&Cube, Sphere, Cylinder, Cone, Torus\\
        \midrule
        Size&Small, Medium, Large\\
        \midrule
        Color&Red, Blue, Yellow, Purple, Orange\\
        \midrule
        Angle&$0^{\circ}, 15^{\circ}, 30^{\circ}, 45^{\circ}, 60^{\circ}, 90^{\circ}$\\
        \bottomrule
        \end{tabular}
        
      \captionof{table}{Data generating factors of CANDLE}
    \label{tab:generating_factors}
	\end{minipage}\hfill
	\begin{minipage}{0.45\linewidth}
		\centering
        \includegraphics[width=0.7\textwidth]{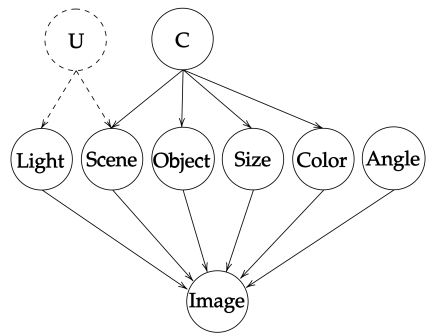}
        
		\captionof{figure}{Data generating mechanism with unobserved(U), observed(C) confounders.}
		\label{fig:datagenerator_confounding_appendix}
	\end{minipage}
\end{table*}

\begin{table*}
\begin{minipage}{0.45\linewidth}
		\centering
        \includegraphics[width=0.65\textwidth]{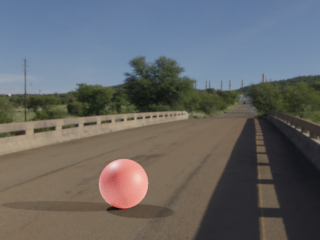}
        
		\captionof{figure}{Sample image taken from CANDLE dataset. On the right: ground truth information about this image in JSON format.}
		\label{fig:image_4150}
	\end{minipage}\hfill
	\begin{minipage}{0.48\linewidth}
		\begin{lstlisting}[firstnumber=1]
{
"scene": "bridge",
"lights": "left",
"objects": {
"Sphere_0": {
    "object_type": "sphere",
    "color": "red",
    "size": 2,
    "rotation": 60,
    "bounds": [[95,29],[154,87]]}}
}
\end{lstlisting}
	\end{minipage}

\end{table*}
\paragraph{Observed Confounding in CANDLE Dataset.}
In order to allow deep generative models to capture confounding, CANDLE introduces observed confounding in the dataset, which provides a layer of complexity that is important for causal analysis in such models. Table \ref{tab:candle_observed_confounding} shows the specific instances of observed confounding present in CANDLE dataset. These choices are made to improve semantic realism of the images.
\label{analysis_on_dataset}
\begin{table*}
    \begin{center}
    \begin{tabular}{ll}
    \toprule
    \textbf{Observed confounding in CANDLE}& \textbf{Reason for the presence of confounding}\\
    \midrule
    Large objects except torus are not present in indoor scene & Large objects except torus occupy excessive vertical space in the\\
                                                                  & indoor scene making it obtrusive in appearance and semantically\\
                                                                  & implausible\\
    \hline
    Large spheres, large cylinders, large cubes are not present& Large spheres, large cylinders, large cubes appear too large \\
    in tunnel, and moonlit grass scenes                         & to be present in tunnel and moonlit grass scenes\\
    \hline
    Large objects are not present in hall scenes              & Large objects occupy too much space in the hall scene making\\
                                                                 &it obtrusive in appearance\\
    \hline
    Small objects are not present in grassland, garage scenes & Small objects appear imperceptibly small in such backgrounds\\
    \hline
    Yellow objects are not present on bridge, city square scenes & Yellow color overlaps with the warm colors of bridge and\\
                                                                 &city square scenes, making the objects near-unresolvable\\
    \hline
    Orange and yellow objects are not present in station,      &Orange and yellow colors overlap with background colors of\\
    dusk city, and playground scenes                             &station, dusk city and playground scenes, making the objects\\
                                                                 &unresolvable\\
    \hline
    Cones are not present in hall, tunnel, and sky walk scenes & The cone's shape uniquely interacts with the light both behind and\\
                                                                & ahead, in these scenes makeing them appear overly shiny and\\ &unrealistic\\
    \hline
    Orange cones are not present on bridge scene              & Light reflections from orange cone on bridge scenes make \\
                                                                & the orange cones too shiny and unrealistic\\
    \hline
    Spheres are not present in skywalk scenes                  & Due to the smooth flooring in skywalk scene and the small \\                                                                  & contact-surface of the sphere, they interact unrealistically\\
    \bottomrule
    \end{tabular}
    \end{center}
    \caption{Observed confounding in CANDLE dataset}
    \label{tab:candle_observed_confounding}
\end{table*}

\paragraph{Dataset Rendering Process.}
The assets and scripts used to render CANDLE dataset are anonymously available at \url{https://github.com/causal-disentanglement/candle-simulator}. Each value of a factor of variation corresponds to a separate \texttt{.blend} file in a hierarchy. For example, \texttt{objects/cube.blend} just contains a cube and \texttt{scenes/indoor.blend} just contains a texture with the HDRI image. Now, each image can be produced by picking one variant from each of the folders, opening in a single Blender instance and rendering it. The above process is automated by using Blender's Python API while rendering the dataset.

\paragraph{Extensibility of CANDLE Dataset.}
\label{appendix:extension}
Since CANDLE is a simulated rendering of 3D objects in a real HDRI background, the dataset itself is easy to extend by adding different variations of each of the factors and rendering a different version suitable for some specific downstream task (examples are given below). As such, care is taken to ensure that extensibility is one of the goals that this dataset satisfies implicitly.

Extending the dataset is done by modifying or replacing the existing assets (e.g., \texttt{.blend} files) in the hierarchy and rerendering. This can be done with minimal knowledge of Blender. For example, replacing the sphere in Figure \ref{fig:image_4150} with a cuboid can be done in the following simple steps:
\begin{itemize}
\setlength\itemsep{-0.1em}
    \item Open \texttt{objects/sphere.blend} in Blender, select the sphere and hit \texttt{x} to remove it
    \item Add a cube by hitting \texttt{shift+A > mesh > cube}. Select a face, click the move tool in the toolbar in the left and drag to get a cuboid.
    \item Rename the sphere to a cuboid in the panel to the right and save. Rename the file and \texttt{object\_type} in the script too.
\end{itemize}
Now, rerendering using the provided script will result in a variant of the dataset with all instances of the sphere containing a cuboid, with the scene, coloring and other factors applied automatically. A similar simple process extends all factors of variation independently. Further details for rendering and conventions followed in the dataset are provided at: \url{https://github.com/causal-disentanglement/candle-simulator}. We hope these details can be leveraged by interested users of the dataset as needed.

\begin{figure*}
\centering
   \includegraphics[width=0.9\textwidth]{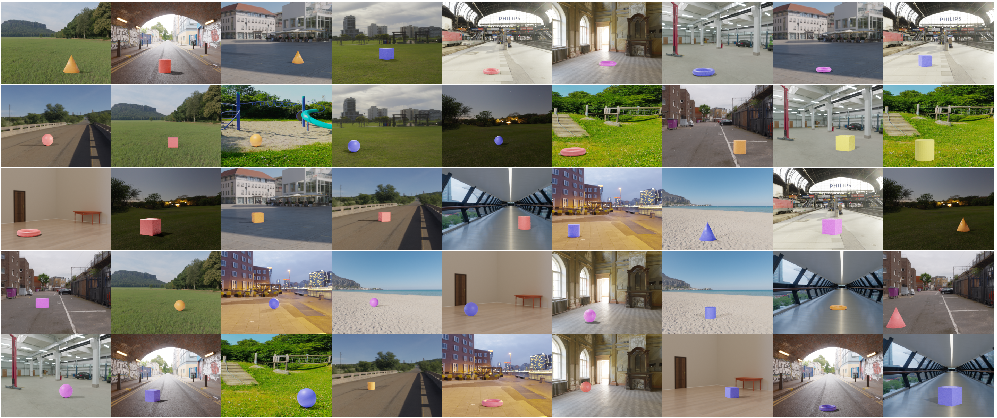}
   \caption{Sample images from the CANDLE dataset}
   \label{appendix:more_samples}
\end{figure*}

\paragraph{More Images from CANDLE Dataset.}
In addition to the images shown in the main paper, Figure \ref{appendix:more_samples} presents some more sample images from the CANDLE dataset. These images demonstrate the dataset's multiple natural backgrounds with simulated objects whose generative properties are known.
\paragraph{Comparison of CANDLE Dataset With Existing Datasets in Disentanglement Literature.}
Figure \ref{appendix:comparison_of_datasets} presents a visual comparison of CANDLE (left) with popular existing datasets (right) in disentanglement literature. CANDLE dataset is the only dataset with a realistic scene and a foreground object controlled by several latent factors among all these datasets. Existing datasets are largely synthetic and/or have a simplistic generative causal process. Table~\ref{tab:datasets} shows the comparison of CANDLE with existing datasets in the disentanglement literature across various dimensions. This comparison suggests that CANDLE dataset is a good choice for studying disentanglement and causal analysis in disentanglement learning.

\section{Additional Experimental Results}
\label{details_on_experiments}
\begin{figure}
	\centering
    \includegraphics[width=0.75\linewidth]{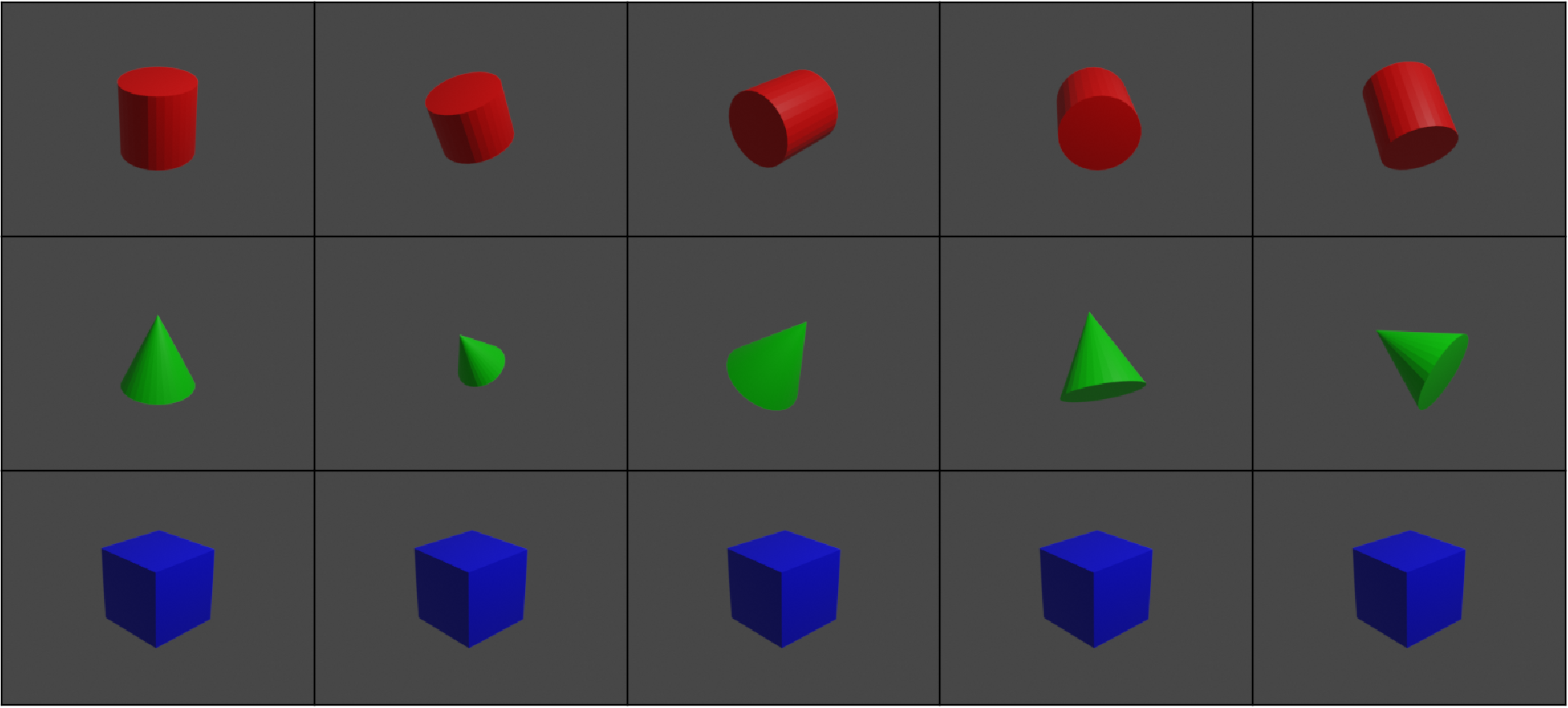}
	\caption{Sample images from synthetic dataset with full observed confounding. Cylinders appear in red, cones in green, and cubes in blue.}
	\label{fig:synthetic_images}
\end{figure}

\begin{figure*}
\centering
   \includegraphics[width=0.8\textwidth]{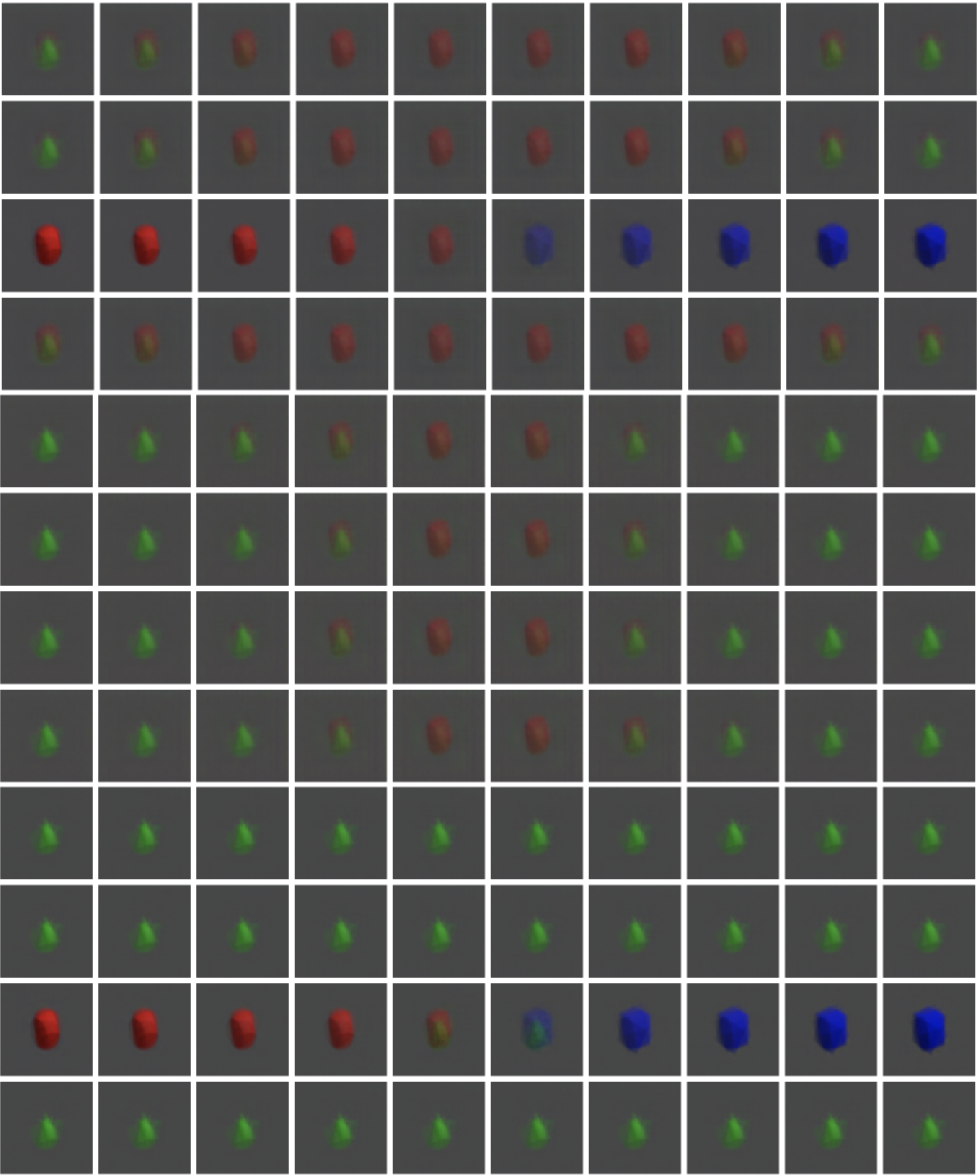}
   \caption{Generated images when a random latent dimension is traversed/interpolated in $\beta$-VAE model on the synthetic dataset. Each row in the above grid shows the generated images when we traverse/interpolate a random latent dimension. It is qualitatively evident from the results that color and shape are confounded by a set of latents. Whenever color changes, shape also changes and vice versa. Usual reconstruction loss is used.}
   \label{appendix:synthetic_images_traverse}
\end{figure*}

\paragraph{Additional Details on Experimental Setup.}
Adding to the details of experimental setup in Section 7 of main paper, in all experiments batch size used is $64$, and latent space dimension is $64$. $\beta$ value for $\beta$-VAE, $\beta$-TC-VAE is $10$ in CANDLE experiments, and $4$ in dSprites and MPI3D-Toy experiments. $\gamma$ value used for Factor-VAE is 4 in CANDLE experiments and 6 in dSprites and MPI3D-Toy experiments. We used DIP-VAE variant 1 (DIP-VAE-I) and its corresponding hyperparameters are $\lambda_d=10$ and $\lambda_{od}=10$ in CANDLE experiments and $\lambda_d=100$ and  $\lambda_{od}=10$ in dSprites and MPI3D-Toy experiments. For semi-supervised methods, the weight for supervised loss is $4$. All experiments and rendering were conducted on a $4\times$ NVIDIA GeForce 1080Ti computing unit.
\paragraph{Pre-trained Classifier Used in CG Metric.}
We use a pre-trained classifier to identify generative factors in an (counterfactual) image. A standard multi-class CNN architecture: (CONV+RELU)x3 + FC is used to predict the value of each generative factor, given an image. For CANDLE, number of output neurons would be sum of all possible values of each generative factor (e.g.,: cube, ..., torus,  red, ..., green, ..., indoor, ..., playground) 
-- 38 in total as in Table~\ref{tab:generating_factors}. We found this CNN architecture to be an easy arbitrary choice, and noticed no significant change in results on changes in architecture.
\paragraph{Experiments on Sythetic Dataset.}
We created a synthetic toy dataset ($432$ images of shape $128\times128$) with full confounding where certain objects appear only in certain colors to assess a model's behavior under such conditions (Fig \ref{fig:synthetic_images}). Here, we consider only two generative factors: \textit{shape} and \textit{color}. Reconstructions and latent traversal of a $\beta$-VAE model trained on the synthetic dataset reveal that both color and shape are captured by the same set of latents (Fig.~\ref{appendix:synthetic_images_traverse}) which is the visual indicator of bad/no disentanglement. Table \ref{tab:synthetic_experiments} shows the quantitative results. Our metrics reveal that all the models perform poorly on the synthetic dataset. The $IRS$ score is however close to $1$, which shows that it may not be suitable for measuring the degree of unconfoundedness achieved by a model. $DCI(D)$ scores are close to $0.1$ but our metrics do an even better job by giving scores of exactly zero ($UC$) and almost zero ($CG$) for the models that fail to disentangle the generative factors under full confounding.

\begin{figure}
    \centering
    \includegraphics[width=0.9\linewidth]{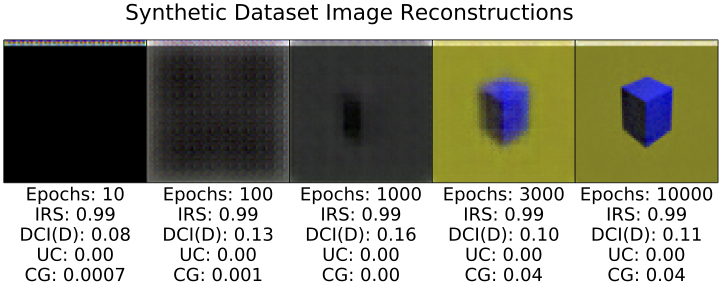}
    \caption{Epochs vs reconstructions of $\beta$-VAE model on synthetic dataset}
    \label{fig:synthetic_comparison}
\end{figure}

We also observed that $IRS$ score is independent of the quality of reconstruction. For example, as shown in Figure~\ref{fig:synthetic_comparison}, even with 10 epochs of training a $\beta$-VAE model, we get IRS score of 0.99 indicating good disentanglement score but with bad reconstructions. The IRS score remains at 0.99 even after getting good reconstructions, which makes it difficult judge the usefulness of IRS score in such datasets. On the other hand, our metrics output the values of exactly zero ($UC=0$, because of confounded latents) and almost zero ($CG\sim 0$, because of model's inability to generate counterfactual images that differ in only one generative factor, which is again expected as $UC=0$).
\begin{table}
	\centering
	\footnotesize
	\begin{tabular}{lcccc}
    \toprule
    Model  &$IRS$&$DCI$&$ UC $& $ CG$\\
    &&($D$)&$\rho=1$&$\rho=1$\\
    \midrule
    $\beta$-VAE &0.99&0.10&0.00&0.01\\
    $\beta$-TCVAE &0.99&0.13&0.00&0.04\\
    DIP-VAE    &0.99&0.11&0.00&0.03\\
    Factor-VAE &0.99&0.12&0.00&0.04\\
    \bottomrule
    \end{tabular}
  \caption{Comparison of $IRS, UC, CG$ metrics on synthetic dataset for various models}
\label{tab:synthetic_experiments}
\end{table}
\paragraph{Confounded dSprites.}
To assess the level of disentanglement under confounding on the dSprites dataset, we performed experiments by selecting images from dSprites according to the conditioning mentioned in Table \ref{tab:confounding_info_dsprites}. This conditional selection mimics the observed confounding as it causes spurious correlations between features. The results are in Table \ref{tab:dsprites_confounding}. Compared to Table 2 of the main paper (where we experimented on dSprites dataset without any confounding), here we observe low $UC$ and $CG$ scores because of the model's inability to perform causal disentanglement in the presence of confounders.
\begin{table}
\begin{center}
    \footnotesize
    \begin{tabular}{lccc}
    \toprule
    Shape & Available  & Available  & Available \\
     &  size &  Orientation &  Position\\
    \midrule
     Square&Small&$0-\frac{2\pi}{3}$ & Top Left\\
     Ellipse&Medium&$\frac{2\pi}{3}-\frac{4\pi}{3}$ & Middle \\
     Heart&Large&$\frac{4\pi}{3}-2\pi$&Bottom Right\\
    \bottomrule
    \end{tabular}
    \end{center}
    \caption{Confounding chosen between object and color in dSprites dataset for experiments in Table \ref{tab:dsprites_confounding}}
    \label{tab:confounding_info_dsprites}
\end{table}
\begin{table}
\begin{center}
\scalebox{0.77}{
\begin{tabular}{lcccccc}
\toprule
Model  & $IRS$&$DCI$ & $UC$&$CG$& $UC$&$CG$\\
 & &$(D)$ & $(\rho=1)$&$(\rho=1)$& $(\rho=2)$&$(\rho=2)$\\
 \midrule
$\beta$-VAE &0.63&0.12&0.63&0.07&\textbf{0.53}&0.07\\
$\beta$-TCVAE  &\textbf{0.75}&0.23&0.33&0.06&0.22&0.07\\
DIP-VAE &0.51&0.10&0.73&\textbf{0.10}&0.40&0.09\\
Factor-VAE  &0.57&0.12&\textbf{0.86}&0.02&0.49&0.03\\
SS-$\beta$-VAE &0.55&0.18&0.73&0.05&0.48&0.05\\
SS-$\beta$-TCVAE &0.70&\textbf{0.36}&0.33&\textbf{0.10}&0.22&\textbf{0.10}\\
SS-DIP-VAE &0.43&0.12&0.80&0.05&0.48&0.05\\
SS-Factor-VAE &0.62&0.25&0.73&0.09&0.48&0.09\\
\bottomrule
\end{tabular}
}
\end{center}
\caption{Comparison of $DCI, IRS, UC$ and $CG$ metrics on dSprites for various models under confounding as given in Table \ref{tab:confounding_info_dsprites}}

\label{tab:dsprites_confounding}
\end{table}

\paragraph{Confounded MPI3D-Toy.}

To assess the level of disentanglement under confounding on the MPI3D-Toy dataset, we performed experiments by selecting images from MPI3D-Toy according to the conditioning mentioned in Table \ref{tab:confounding_info_mpi3d}, and the results are in Table \ref{tab:mpi3d_confounding}. Compared to Table 3 of the main paper (where we experimented on MPI3D-Toy dataset without any confounding), here too we observe low $UC$ and $CG$ scores because of the model's inability to perform causal disentanglement under confounding.
\begin{table}
    \centering
    \footnotesize
    \begin{tabular}{ccccc}
    \toprule
    Color&Shape&Size&h-axis&v-axis\\
    \midrule
    Green&Cube, Cylinder&Small&0-10&0-10\\
    Red&Cylinder, Sphere&Large&10-20&10-20\\
    Blue&Sphere, Cube&Small, Large&20-30&20-30\\
     \bottomrule
    \end{tabular}
    \caption{Confounding chosen between object and color in MPI-3D dataset for experiments in Table \ref{tab:mpi3d_confounding}; all images are centered with \textit{height=1}, \textit{background\ color} assumes all possible values.}
    \label{tab:confounding_info_mpi3d}
\end{table}

\begin{table}
\begin{center}
\scalebox{0.77}{
\begin{tabular}{lcccccc}
\toprule
Model  & $IRS$&$DCI$ & $UC$&$CG$& $UC$&$CG$\\
  & &$(D)$ & $(\rho=1)$&$(\rho=1)$& $(\rho=2)$&$(\rho=2)$\\
\midrule
$\beta$-VAE &0.18&\textbf{0.01}&$0.28$&0.11&0.19&0.11\\
$\beta$-TCVAE  &0.38&0.008&0.00&0.10&0.00&\textbf{0.21}\\
DIP-VAE &0.12&0.005&0.28&0.06&0.19&0.08\\
Factor-VAE  &0.26&\textbf{0.01}&0.66&\textbf{0.14}&\textbf{0.38}&0.13\\
SS-$\beta$-VAE &0.63&0.006&0.66&0.12&0.19&\textbf{0.21}\\
SS-$\beta$-TCVAE &\textbf{0.64}&0.007&0.28&0.06&0.19&0.12\\
SS-DIP-VAE &0.32&0.007&\textbf{0.76}&0.06&0.32&0.10\\
SS-Factor-VAE &0.50&0.006&0.00&0.12&0.00&0.20\\
\bottomrule
\end{tabular}
}
\end{center}
\caption{Comparison of $DCI, IRS, UC$ and $CG$ metrics on MPI3D for various models under confounding as given in Table \ref{tab:confounding_info_mpi3d}}

\label{tab:mpi3d_confounding}
\end{table}

\begin{figure*}
\centering
   \includegraphics[width=0.6\textwidth]{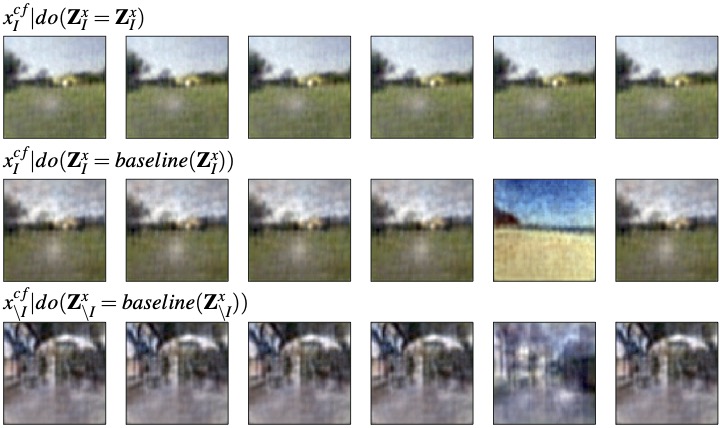}
   \caption{Counterfactual images obtained as part of studies on $CG$ metric($\rho=5$) for proposed SS-Factor-VAE model on CANDLE. The white spot on these images correspond to the foreground object and is not clearly captured because of model's inability to capture moving foreground objects in a complex background. Top row shows original reconstructions (no change in latents). Middle row shows  counterfactual images generated when latent dimensions corresponding to six generative factors are set to baseline values. Bottom row shows counterfactual images generated when latent dimensions not corresponding to the six generative factors are set to baseline values. Each column corresponds to change in latent dimensions corresponding to generative factors: shape, color, size, rotation, scene, and light respectively. All columns except column 5 have similar looking images because those images are generated by changing the latents corresponding to change in latent dimensions of foreground object's properties (size, shape etc) which are not clearly captured by the models and this behavior is observed across state-of-the-art generative disentanglement models.}
   \label{appendix:counterfactuals_ss_factor_vae}
\end{figure*}

\begin{figure*}
\centering
   \includegraphics[width=0.6\textwidth]{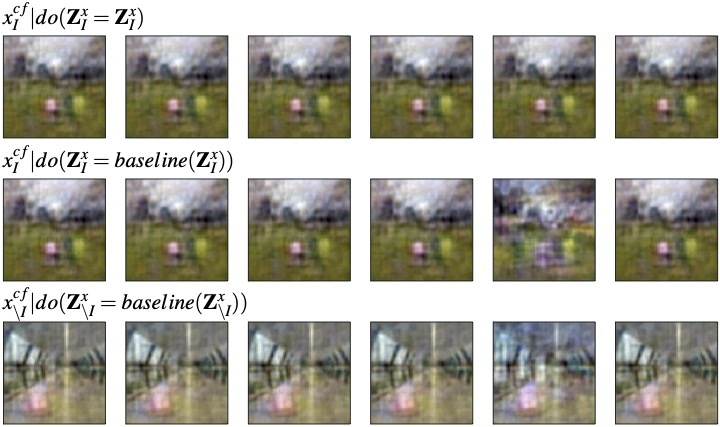}
   \caption{Counterfactual images obtained as part of studies on $CG$ metric ($\rho=5$) for SS-Factor-VAE-BB model on CANDLE. Foreground objects are retained but in this case, more than one foreground object appears in the images. This motivates the need for further research in causal disentanglement with focus on specific objects in a given image.}
   
   \label{appendix:counterfactuals_ss_fvae_vae_cg}
\end{figure*}

\paragraph{Counterfactual Images Generated for $CG$ Computation.}
Figures \ref{appendix:counterfactuals_ss_factor_vae},\ref{appendix:counterfactuals_ss_fvae_vae_cg} show the counterfactual images generated while calculating the $CG$ metric. We note that well-known state-of-the-art models are unable to capture the foreground across the counterfactuals. This is because of both the confounding effect as well as limitations of capturing small/moving foreground objects with complex backgrounds. Our proposed method \textit{SS-FVAE-BB} however retains foreground objects during counterfactual generation with some limitations as explained below in the figures. Our dataset thus helps assess existing disentanglement models on how they respond to confounding generative factors/latent variables, and motivates the development of better models for this purpose. 


\begin{figure*}
   \includegraphics[width=1\textwidth]{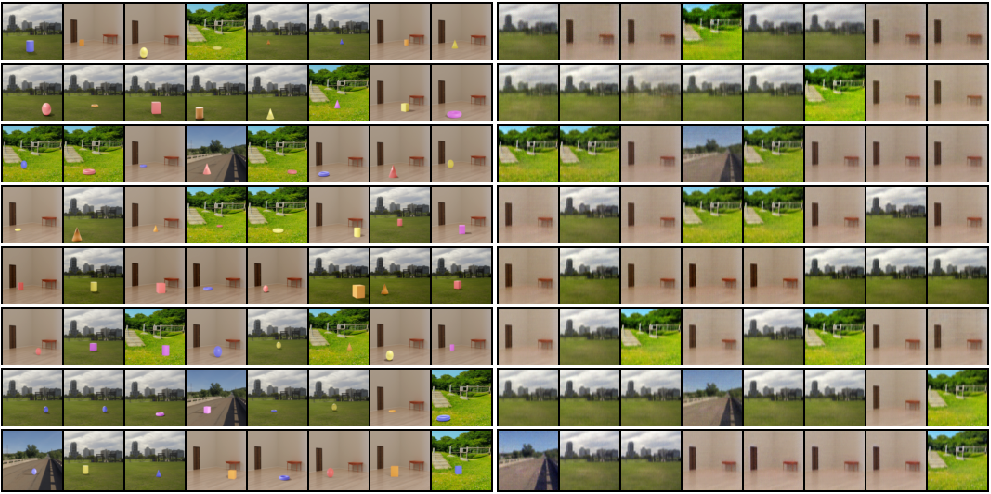}
   \caption{\textit{(Best viewed in color, zoomed in)} Left grid contains original images from CANDLE and right grid shows the reconstructions of those images by $\beta$-VAE model with usual reconstruction loss. Here $\beta$-VAE model is failed to capture foreground objects. A similar phenomenon is observed in other existing disentanglement methods as well which suggests the need for better disentanglement methods and CANDLE is a good choice to study such methods.}
   \label{fig:recon_beta}
\end{figure*}

\begin{figure*}
   \includegraphics[width=1\textwidth]{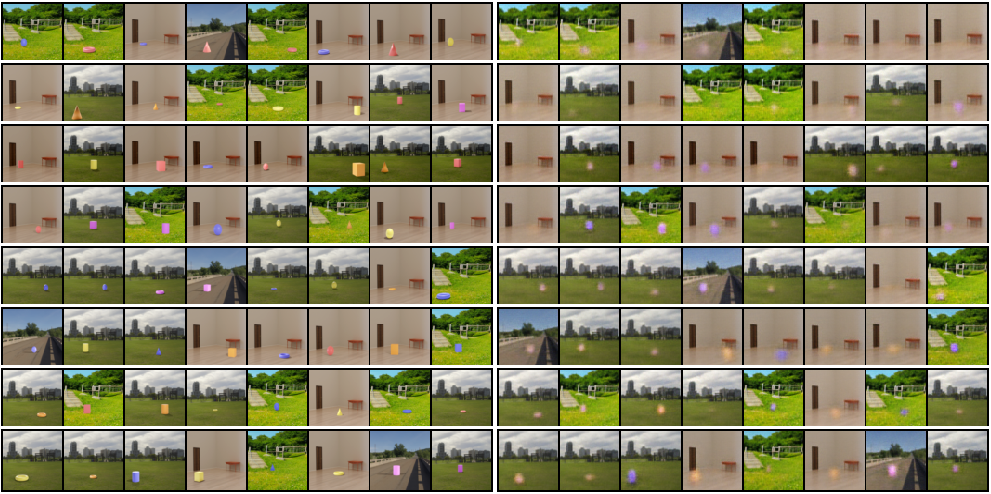}
   \caption{\textit{(Best viewed in color, zoomed in)} Left grid contains original images from CANDLE and right grid shows the reconstructions of those images by $\beta$-VAE model. Unlike for the experiments shown in Figure~\ref{fig:recon_beta}, this time reconstruction loss term is scaled by a factor of 3000. Because of this large multiplicative factor, objects are retained better in reconstructions but due to a relatively lesser weight for the KL-divergence loss term, latent representations are not guaranteed to be learned well~\cite{kim2018disentangling}. This again shows the need for better disentanglement methods to work on datasets such as CANDLE. Recall, we partially solved this problem in Section 6 of main paper using bounding box supervision.}
   \label{fig:recon}
\end{figure*}

\paragraph{Qualitative Results of Experiments on CANDLE and Synthetic Datasets.}
Figures \ref{fig:recon_beta}-\ref{appendix:synthetic_images_traverse} present some additional qualitative results as part of ablation studies on CANDLE and Synthetic datasets.

\section{Assets and Licensing}
The assets created in this work, namely the CANDLE dataset itself, is available at \url{https://causal-disentanglement.github.io/IITH-CANDLE/} under the Creative Commons Attribution 4.0 International License. The anonymized code used to reproduce the dataset can be found at \url{https://github.com/causal-disentanglement/candle-simulator} under the MIT license and the code to reproduce experimental results can be found at \url{https://github.com/causal-disentanglement/disentanglement_lib} under the Apache License 2.0. Specifically, the HDRI images used as backgrounds in the dataset's creation are publicly available under a CC0 license. To the best of our knowledge, the assets, libraries and tools used are open-source and have been cited.
Instructions to reproduce the experiments and the dataset are provided in the anonymized code repository itself as well as in Section \ref{appendix:extension}.
\end{document}